\documentclass{article}

\PassOptionsToPackage{numbers,compress}{natbib}
\usepackage[preprint]{neurips_2026}

\usepackage[utf8]{inputenc} 
\usepackage[T1]{fontenc}    
\usepackage{hyperref}       
\usepackage{url}            
\usepackage{booktabs}       
\usepackage{amsfonts}       
\usepackage{nicefrac}       
\usepackage{microtype}      
\usepackage{xcolor}         
\usepackage{graphicx}
\usepackage{caption}
\usepackage{wrapfig}
\usepackage{amsmath}
\usepackage{multirow}
\usepackage{comment}
\usepackage{bm}

\title{Overcoming Kernel Redundancy for\\ Scaling Logic Gate Networks}

\author{
  Sejin Park \quad\quad
  Hongjae Lee \quad\quad
  Changwoo Han \quad\quad
  Seung-Won Jung\thanks{Corresponding author.} \\
  \\
  Department of Electrical Engineering \\
  Korea University \\
  \texttt{\{sejin5485, jimmy9704, hcwoo329, swjung83\}@korea.ac.kr}
}

\begin{document}

\maketitle



\begin{abstract}
Differentiable logic gate networks, which operate using only logic gates, have recently attracted attention as an efficient alternative to conventional neural networks. However, despite their efficiency, the scaling behavior of logic gate networks remains underexplored. By contrast, scaling model capacity is a central design principle in deep neural networks and typically leads to improved performance. This discrepancy raises a key question: Can similar scaling benefits also be achieved in logic gate networks? In this work, we focus on width as a primary scaling axis and conduct a systematic analysis of its behavior in logic gate networks. We observe that naive width scaling often introduces redundancy among logic kernels, limiting the effective use of additional kernels and leading to performance saturation. To address this limitation, we propose a dynamic logic kernel framework that reorganizes kernel utilization by promoting specialization across kernel groups. This enables the network to better utilize increased width via input-dependent kernel routing, while ensuring that both routing and computation are implemented entirely with gate-level Boolean operations at inference time. We further find that kernel redundancy is most pronounced at the first gate level, motivating an early-stage dynamic logic kernel strategy that concentrates adaptation at this level. Experimental results demonstrate that our approach improves kernel utilization and increases kernel diversity, leading to higher accuracy with improved parameter efficiency.

\end{abstract}

\section{Introduction}

Deep learning has shown remarkable performance across a wide range of tasks; however, its substantial computational cost has driven increasing interest in efficient model design~\citep{howard2017mobilenets, tan2019efficientnet}. To address this challenge, various approaches have been proposed to reduce computational cost while maintaining performance. For example, quantization lowers numerical precision to reduce computation while preserving most of the accuracy~\citep{jacob2018quantization, zhou2016dorefa}, and binary neural networks further simplify computation by representing both weights and activations with binary values, enabling efficient bit-wise operations~\citep{hubara2016binarized, rastegari2016xnor}. These efficiency-oriented approaches have been successfully applied across diverse domains, including image restoration, language modeling, and other real-world applications~\citep{liu2024qllm, park2025lut}.

Recently, differentiable logic gate networks have emerged as a promising direction for efficient deep learning, where computations are performed using only logic gate operations instead of conventional arithmetic operations~\citep{petersen2022deep}. By leveraging Boolean functions and bit-level representations, these models enable highly efficient and lightweight inference, making them well-suited for resource-constrained environments. Building upon this paradigm, logic gate networks have been extended to structured architectures such as logic tree-based models, where localized tree kernels operate analogously to convolutional layers~\citep{petersen2024convolutional}. In parallel, recent studies have focused on improving the optimization of logic-based models through methods such as continuous relaxation and reparameterization~\citep{yousefi2026mind, ruttgers2026light}.

Despite these advances, existing works primarily focus on improving training strategies or architectural design, with limited attention to how model capacity is structured and utilized under scaling~\citep{bello2021revisiting, ghorbani2022scaling}. In particular, while increasing model width introduces multiple logic kernels, it remains unclear whether this additional capacity leads to meaningful functional diversity or merely redundant representations. In contrast to modern deep neural networks, where increased model capacity typically translates into consistent performance gains, such scaling behavior is not yet established in logic gate networks. This raises the question of how multiple kernels are utilized in practice and how they contribute to performance as the model scales.

To answer this question, we conduct a systematic analysis of width scaling in logic gate networks. Our analysis reveals that increasing the number of logic kernels through width scaling leads to redundant kernels, resulting in performance saturation. This finding suggests that the limitation of width scaling does not stem from insufficient capacity, but rather from inefficient utilization of the available kernels.


Based on this insight, we propose Dynamic Logic Kernel (DLK), a logic gate-based routing framework that selects kernel groups according to each input. Rather than using a conventional real-valued gating module, DLK formulates kernel group selection itself as a Boolean routing operation, keeping routing computation within the logic gate domain at inference time. By specializing different kernel groups for different inputs, DLK allows additional width to be used more effectively. We further examine where redundancy appears within the logic kernel hierarchy and find that redundancy is most pronounced at the first gate level. This motivates Early-stage Dynamic Logic Kernel (EDLK), a logic gate network-specific dynamic mechanism that routes first-level gate groups as logic components.

Our contributions are as follows:

\begin{itemize}
    \item \textbf{Scaling analysis.} 
    We present a systematic analysis of width scaling in logic gate networks and reveal that naively increasing the number of logic kernels induces redundancy, resulting in inefficient kernel utilization and performance saturation.

    \item \textbf{Logic kernel specialization.} 
    We propose Dynamic Logic Kernel (DLK), a logic gate-based
    routing mechanism that reorganizes kernel utilization through input-dependent group selection. This promotes specialization across kernel groups while ensuring that both routing and computation are implemented entirely using gate-level Boolean operations.

    \item \textbf{Early-stage adaptation.} 
     Through level-wise kernel analysis, we identify that redundancy is most pronounced at the first gate level. This observation motivates Early-stage Dynamic Logic Kernel (EDLK), which introduces a logic gate network-specific dynamic mechanism that treats first-level gate groups as selectively routable logic components.

    \item \textbf{Scalable efficiency and performance.} 
    We demonstrate that our approach improves kernel diversity and model-capacity utilization, leading to improved performance under width scaling.
\end{itemize}

\section{Background}
\subsection{Preliminaries}

\paragraph{Differentiable logic gate networks.}
Logic gate networks replace conventional neurons with Boolean operators such as AND, OR, and XOR, where each neuron selects one of 16 possible Boolean functions. Since Boolean functions are discrete, differentiable logic gate networks use continuous relaxations of logic operators~\citep{petersen2022deep}. Following prior work, let $g_1,\dots,g_{16}$ denote the 16 relaxed binary logic operators. Then, for a trainable parameter vector ${\mathbf{z}} \in \mathbb{R}^{16}$, a differentiable logic gate $f_{\mathbf{z}}$ is defined as
\begin{equation}
f_{\bf{z}}(a_1,a_2)=\sum_{i=1}^{16}\frac{e^{{\mathbf{z}}_i}}{\sum_{j=1}^{16} e^{{\mathbf{z}}_j}}\, g_i(a_1,a_2),
\end{equation}
where ${\mathbf{z}}_i$ denotes the $i$-th element of ${\mathbf{z}}$, and $a_1,a_2 \in [0,1]$ denote the inputs to the gate. After training, the gate is discretized by selecting the operator with the highest probability.

To improve expressivity, multiple logic gates are composed into a tree structure that serves as a kernel, referred to as a logic tree~\citep{petersen2024convolutional}. A logic kernel applies logic gates hierarchically, where each internal node takes two inputs and produces one output. For example, a depth-two logic kernel can be represented as $
f_{\mathbf{z}_3}\big(f_{\mathbf{z}_1}(a_1,a_2),\, f_{\mathbf{z}_2}(a_3,a_4)\big)$. Such logic kernels are applied over local receptive fields and operate analogously to convolutional filters, where multiple kernels are used in parallel to extract features from different input regions. In this formulation, increasing the number of kernels corresponds to width scaling of the network. For more details about logic gate networks, please refer to prior work~\citep{petersen2024convolutional}.

\subsection{Related Work}

Recent studies have explored several directions for improving logic gate networks, including discrepancy reduction and reparameterization. For example, stochastic relaxations based on Gumbel-Softmax have been introduced to alleviate the discretization gap between continuous training and discrete inference~\cite{yousefi2026mind}. Another line of work reformulates the parameterization of logic gate networks using truth-table representations~\cite{ruttgers2026light}. These approaches improve optimization stability and reduce discretization errors through more effective parameterization and relaxation strategies. 

In addition, logic gate networks have also been extended to broader application domains. LogicCBM introduces differentiable logic operations into concept-based models to capture non-linear relationships between concepts beyond linear aggregation~\citep{vemuri2026logiccbms}. Differentiable logic circuits have also been proposed for image compression, replacing conventional neural operations with trainable logic or lookup-based computations for hardware-friendly inference~\citep{aczel2026ic}. More recently, LogicIR extends logic gate networks to image restoration, introducing a fully logic-gate architecture for low-level vision tasks such as image denoising, deblocking, and deraining~\citep{lee2026logicir}.

Despite these advances, a fundamental understanding of logic gate networks remains limited. In particular, the role of width scaling has not been systematically studied, and it remains unclear how logic kernels are utilized within the network. To address this gap, Section~\ref{sec:motivation} systematically analyzes kernel utilization under width scaling and reveals redundancy among logic kernels. Building on this analysis, Section~\ref{sec:method} introduces the proposed method in two stages: Dynamic Logic Kernel (DLK) and Early-stage Dynamic Logic Kernel (EDLK), the latter motivated by hierarchical diversity analysis.

\section{Scaling Behavior in Logic Gate Networks}
\label{sec:motivation}
In conventional deep neural networks, scaling model capacity through width or depth is generally expected to yield consistent performance improvements~\citep{ghorbani2022scaling}. However, such scaling behavior is not directly transferable to logic gate networks. Previous work has shown that simply increasing the depth of differentiable logic gate networks does not necessarily yield consistent performance improvements; instead, deeper models can introduce optimization and discretization challenges, such as vanishing gradients and degraded discretized accuracy~\citep{yousefi2026mind,ruttgers2026light}.

This suggests that scaling strategies in logic gate networks may differ fundamentally from those for conventional neural networks. More importantly, the role of width scaling in logic gate networks remains largely underexplored. This raises the question of whether increasing model width improves performance or encounters limitations similar to those observed in depth scaling.

\subsection{Effect of Width Scaling}
\label{sec:width_scaling}

\begin{wrapfigure}{r}{0.47\linewidth}
    \centering
    \vspace{-45pt}
    \includegraphics[width=1.0\linewidth]{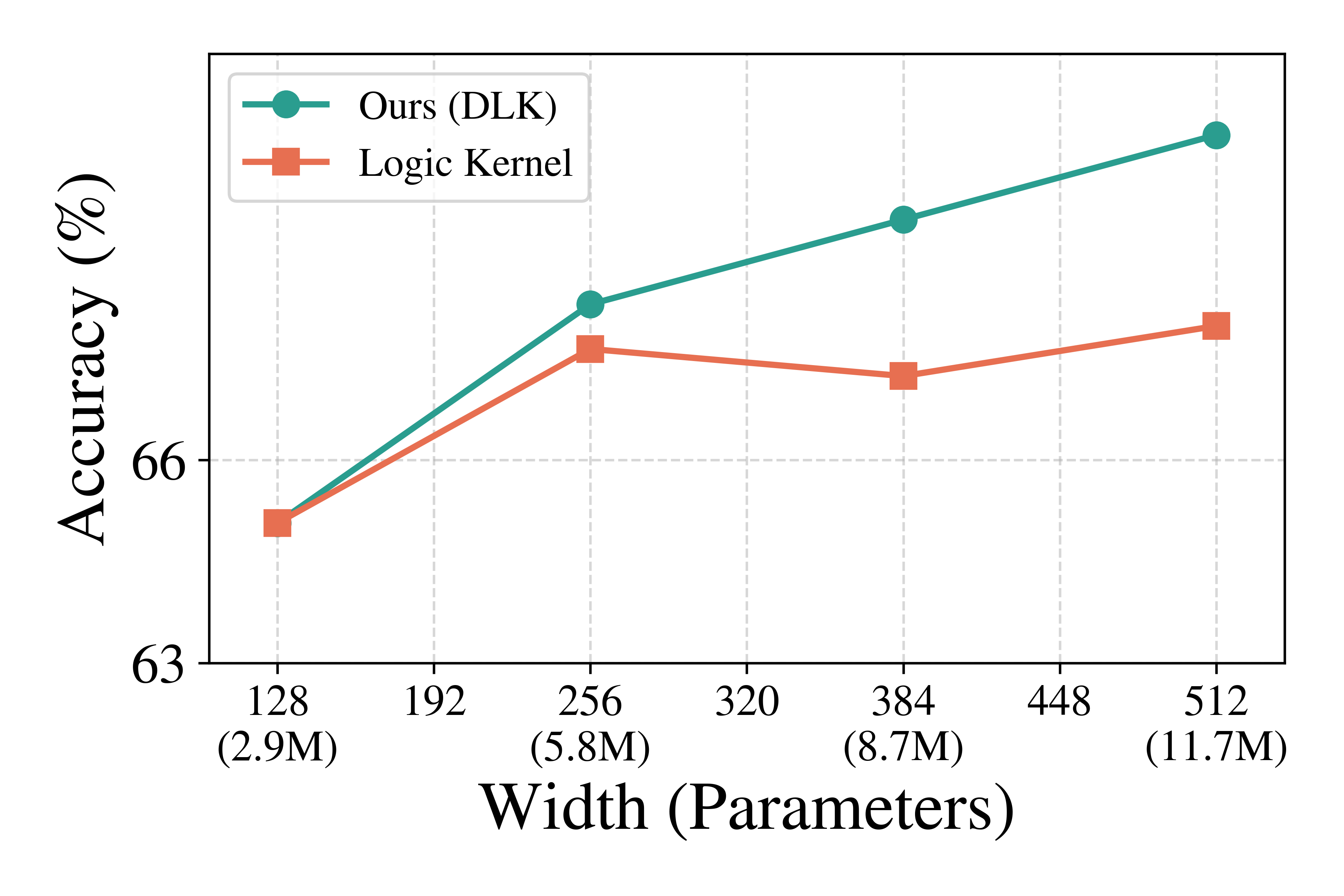}
    \caption{Accuracy under width scaling on the CIFAR-10 dataset.}
    \label{fig:width-scaling}
    \vspace{-20pt}
\end{wrapfigure}

To investigate this question, we first evaluate the effect of width scaling by increasing the number of logic kernels. We use the same model architecture and training settings, varying only the width setting.

Figure~\ref{fig:width-scaling} summarizes performance under different widths. Here, ``Width'' denotes the base width setting---the number of logic kernels in the first convolutional logic layer---from which subsequent convolutional layers are scaled by fixed multipliers; ``Parameters'' reports the total model parameters.

As shown in Figure~\ref{fig:width-scaling}, the standard ``Logic Kernel'' baseline exhibits performance saturation, where increasing the number of kernels does not lead to consistent performance improvements. This observation suggests that simply increasing width is insufficient to effectively utilize the additional capacity. To understand this limitation, we next analyze why naive width scaling fails to improve performance in logic gate networks.

\subsection{Diversity of Logic Kernels}
\label{sec:understanding_logic_tree}

Motivated by the above observation, we hypothesize that the performance saturation arises from redundancy among logic kernels. 
In particular, increasing the number of kernels may cause many of them to produce similar responses rather than capture diverse input patterns. Such redundancy could limit the extent of performance improvement. To examine this hypothesis, we analyze the diversity of logic kernels.

\paragraph{Logic kernel diversity metric.}

Tree edit distance (TED) has been widely used to measure similarity between tree-structured objects. It quantifies the minimum cost required to transform one tree into another through a sequence of edit operations~\citep{zhang1992editing}. In this work, we consider the case where two logic trees share the same topology, under which TED simplifies to a comparison of gate operators at corresponding nodes. Given two trees, $T_A$ and $T_B$, with identical topology, their TED can be defined as the number of differing gate operators at corresponding nodes:
\begin{equation}
d(T_A,T_B)=\sum_{n=1}^{L} \delta_s\!\left(g^{T_A(n)}, g^{T_B(n)}\right),
\quad
\delta_s\!\left(g^{T_A(n)}, g^{T_B(n)}\right)=
\begin{cases}
0, & \text{if } g^{T_A(n)} = g^{T_B(n)},\\
1, & \text{otherwise},
\end{cases}
\label{eq:deltas}
\end{equation}
where $g^{T_A(n)}$ denotes the gate at the $n$-th node of tree $T_A$, and $L$ is the number of nodes.

\begin{wrapfigure}{r}{0.4\linewidth}
    \centering
    \vspace{-15pt}
    \includegraphics[width=\linewidth]{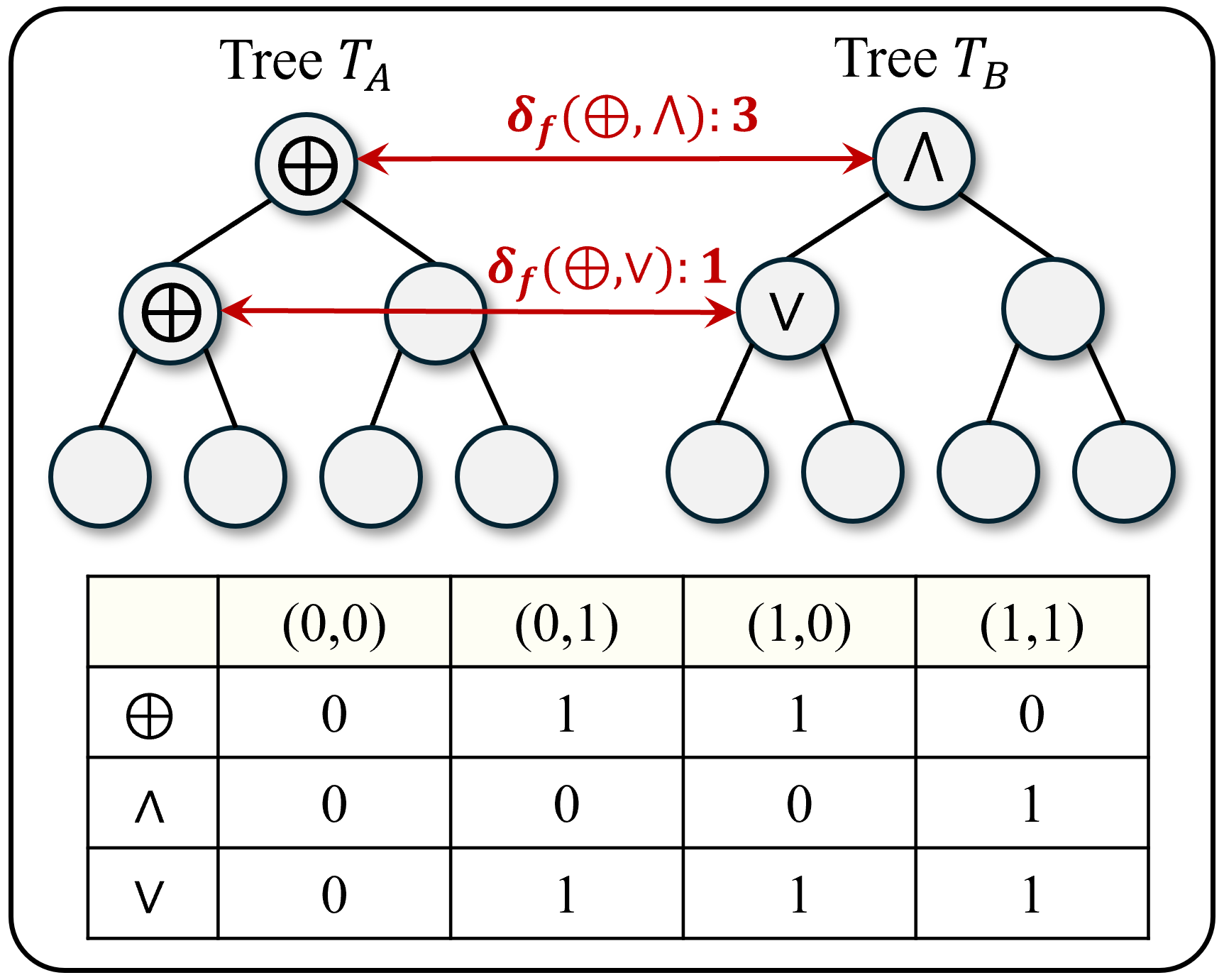}
    \caption{
    Function-aware distance between two aligned logic gates.
    }
    \label{fig:ted_overview}
    \vspace{-10pt}
\end{wrapfigure}
Each gate in a logic tree corresponds to a Boolean function defined by its outputs over a set of input pairs $\mathcal{A} =\{(0,0),(0,1),(1,0),(1,1)\}$. However, this simplified TED formulation treats all gate mismatches equally and does not account for their functional behavior~\citep{zhang1992editing}. It therefore cannot distinguish gates with similar input-output behavior from those with substantially different behavior, limiting its ability to capture meaningful functional differences.


For example, the Boolean function $\oplus$ differs from $\vee$ in only one input case, whereas it differs from $\wedge$ in three input cases. This observation highlights the need for a function-aware distance metric. Accordingly, we define the distance between two gates as 
\begin{equation}
\delta_f(g, g') = \sum_{(a_1,a_2)\in\mathcal{A}} \mathbf{1}\!\left(g(a_1,a_2) \neq g'(a_1,a_2)\right),
\end{equation}
where $\mathbf{1}(\cdot)$ denotes the indicator function, $g(a_1,a_2)$ denotes the output of the Boolean function associated with gate $g$, and the summation counts the number of mismatched outputs over all input combinations. As illustrated in Figure~\ref{fig:ted_overview}, $\delta_f(\oplus, \vee) = 1$ and $\delta_f(\oplus, \wedge) = 3$. By replacing $\delta_s$ with $\delta_f$ in Eq.~\eqref{eq:deltas}, we measure the functional diversity within each convolutional logic layer. Given the set of $N$ logic kernels in a layer, $\mathcal{T}=\{T_1, \dots, T_N\}$, we compute all pairwise distances and summarize the diversity of the kernel set $\mathcal{T}$, denoted as ${\mathcal{D}_\mathcal{T}}$, by averaging these distances over all kernel pairs:
\begin{equation}
{\mathcal{D}_\mathcal{T}}=\frac{1}{\binom{N}{2}} \sum_{i<j} d(T_i, T_j).
\end{equation}
The final diversity is obtained by averaging ${\mathcal{D}_\mathcal{T}}$ across all convolutional logic layers.

\paragraph{Analysis of the kernel diversity.}

\begin{wraptable}{r}{0.45\linewidth}
    \centering
    \small
    \vspace{-15pt}
    \caption{Kernel diversity across widths.}
    \vspace{-3pt}
    \label{tab:TED_width}
    \begin{tabular}{c|cccc}
        \toprule
        Method & 128 & 256 & 384 & 512 \\
        \midrule
        Logic Kernel & 3.05 & 2.50 & 2.05 & 1.86 \\
        Ours (DLK)   & 3.05 & 4.61 & 4.78 & 4.89 \\
        \bottomrule
    \end{tabular}
    \vspace{-15pt}
\end{wraptable}


Table~\ref{tab:TED_width} reports kernel diversity measured using our function-aware TED-based metric, which summarizes truth-table differences between aligned Boolean gates in logic kernels. For the conventional method, diversity remains low and further decreases as width increases, indicating that additional kernels become increasingly redundant. This trend is consistent with the performance saturation observed under width scaling.



Overall, these results suggest that naive width scaling tends to introduce redundant kernels rather than translate additional width into effective capacity. This motivates the dynamic logic kernel framework introduced in the next section, which aims to improve kernel utilization under width scaling.



\section{Dynamic Kernel Utilization for Scaling Logic Gate Networks}
\label{sec:method}

As discussed in Section~\ref{sec:motivation}, width scaling in logic gate networks is limited by redundancy among logic kernels rather than by a lack of capacity. 
This observation shifts the problem from increasing the number of kernels to how they are utilized. In particular, increasing width is most effective when the additional logic kernels produce informative and non-redundant responses. Related ideas have been explored in CNNs through dynamic filter or branch selection, where filters or computational branches are adapted to each input~\citep{jia2016dynamic,li2019selective,yang2019condconv,chen2020dynamic}. Appendix~\ref{app:cnn_dynamic_kernel} provides further discussion of these prior studies and their relation to our logic kernel selection framework. 


Therefore, to address this limitation, we propose Dynamic Logic Kernel (DLK), a framework that reorganizes logic kernel utilization through input-dependent selection of logic kernels within a fully logic gate-based formulation. In this work, we treat routing as a logical operation, avoiding real-valued gating functions and continuous mechanisms at inference time. This ensures that the selection process remains fully discrete and consistent with the logic gate formulation. It also allows the additional kernels introduced by width scaling to serve as adaptive candidate groups instead of being applied uniformly. We further examine where redundancy appears within the logic kernel hierarchy. This analysis motivates Early-stage Dynamic Logic Kernel (EDLK), which applies dynamic routing only to first-level gate groups by exploiting the Boolean gate hierarchy of logic kernels.



\begin{figure}[t]
    \centering
    \includegraphics[width=\linewidth]{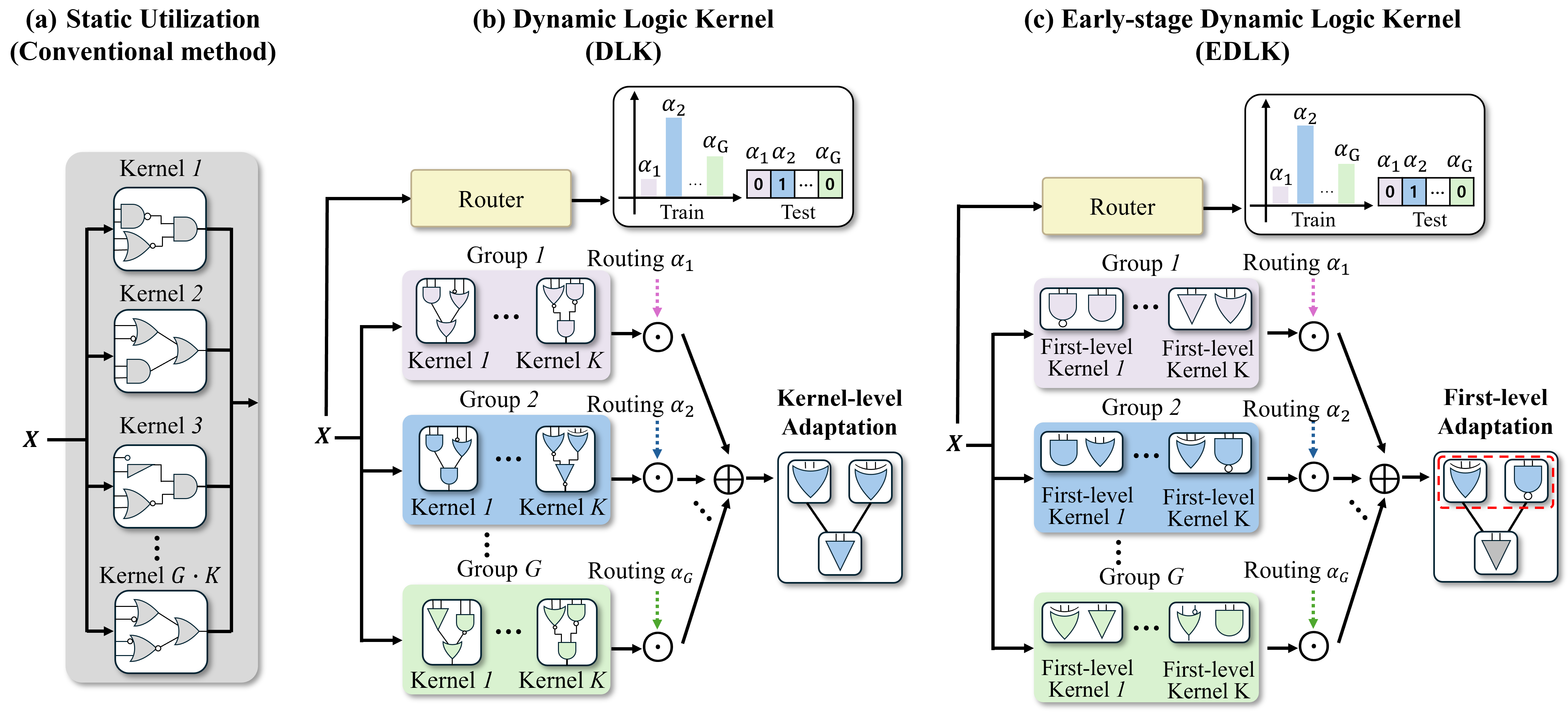}

    \caption{
Overview of kernel utilization strategies in logic gate networks: (a) Static utilization, where all kernels are applied without any selection mechanism, (b) dynamic logic kernel (Sec.~\ref{sec:Selective}), 
(c) early-stage dynamic logic kernel (Sec.~\ref{sec:edlk}).
}
    \label{fig:method_overview}
\end{figure}

\subsection{Dynamic Logic Kernels}
\label{sec:Selective}
\paragraph{Method.} The proposed method is a general module that can be applied to any convolutional logic layer in the network while maintaining a fully logic gate-based formulation. As illustrated in Figure~\ref{fig:method_overview}(b), the method dynamically routes each input to a selected kernel group, thereby improving kernel utilization. For clarity of explanation, we describe the method using an intermediate convolutional logic layer as an example.

The input feature $\bm{X}$ is fed to $G$ kernel groups, each consisting of $K$ logic kernels. 
Let $T_{g,k}(\cdot)$ denote the $k$-th logic kernel in the $g$-th group, where $k = 1, 2, \dots, K$. 
The $g$-th kernel group is defined by concatenating its $K$ kernels as
\begin{equation}
\mathcal{T}_g(\cdot) = \big[ T_{g,1}(\cdot) \; ; \; T_{g,2}(\cdot) \; ; \; \cdots \; ; \; T_{g,K}(\cdot) \big], 
\quad g = 1, 2, \dots, G,
\end{equation}
where $[;]$ denotes concatenation of kernels within the group. Accordingly, the output of each group is given by
\begin{equation}
\mathcal{T}_g(\bm{X}) \in \mathbb{R}^{K \times H_1 \times W_1}.
\end{equation}

To adapt kernel utilization to each input, we introduce a logic router that produces Boolean routing evidence for each kernel group and selectively activates kernel groups that can extract more informative features from the input. The logic router is implemented by convolutional logic layers followed by global OR-pooling over the spatial dimensions.

Specifically, the router $\mathcal{R}(\cdot)$ takes $\bm{X}$ as input and produces $\bm{R} \in [0,1]^{G \times C}$ as follows:
\begin{equation}
\bm{R} = \mathcal{R}(\bm{X}),
\end{equation}
where $C$ denotes the number of bits per group. To obtain a selection score for each group, we aggregate routing bits $\bm{R}$ along the bit dimension as $s_g = \sum_{c} \bm{R}_{g,c}$ for $g=1,\dots,G$. This score can be interpreted as the amount of Boolean evidence supporting each kernel group. During training, the routing weight $\alpha_g$ is computed by applying a softmax with temperature $\tau$ over the selection scores of all groups:
\begin{equation}
\alpha_g =
\frac{e^{s_g/\tau}}
{\sum_{g'=1}^{G} e^{s_{g'}/\tau}}.
\label{eq:alpha}
\end{equation}
The output feature $\bm{Y}$ is then given by
\begin{equation}
\bm{Y} = \sum_{g=1}^{G} \alpha_g \mathcal{T}_g(\bm{X}).
\end{equation}

At inference time, the routing scores $s_g$ are computed using a bit-count operation, and we perform hard group selection as
\begin{equation}
g^* = \arg\max_g {s}_g,
\label{eq:hard_selection}
\end{equation}
and the output is given by
\begin{equation}
\bm{Y} = \mathcal{T}_{g^*}(\bm{X}).
\end{equation}


\label{sec:Early_Selective}
\begin{wrapfigure}{r}{0.38\linewidth}
    \centering
    \vspace{-23pt}
    \includegraphics[width=\linewidth]{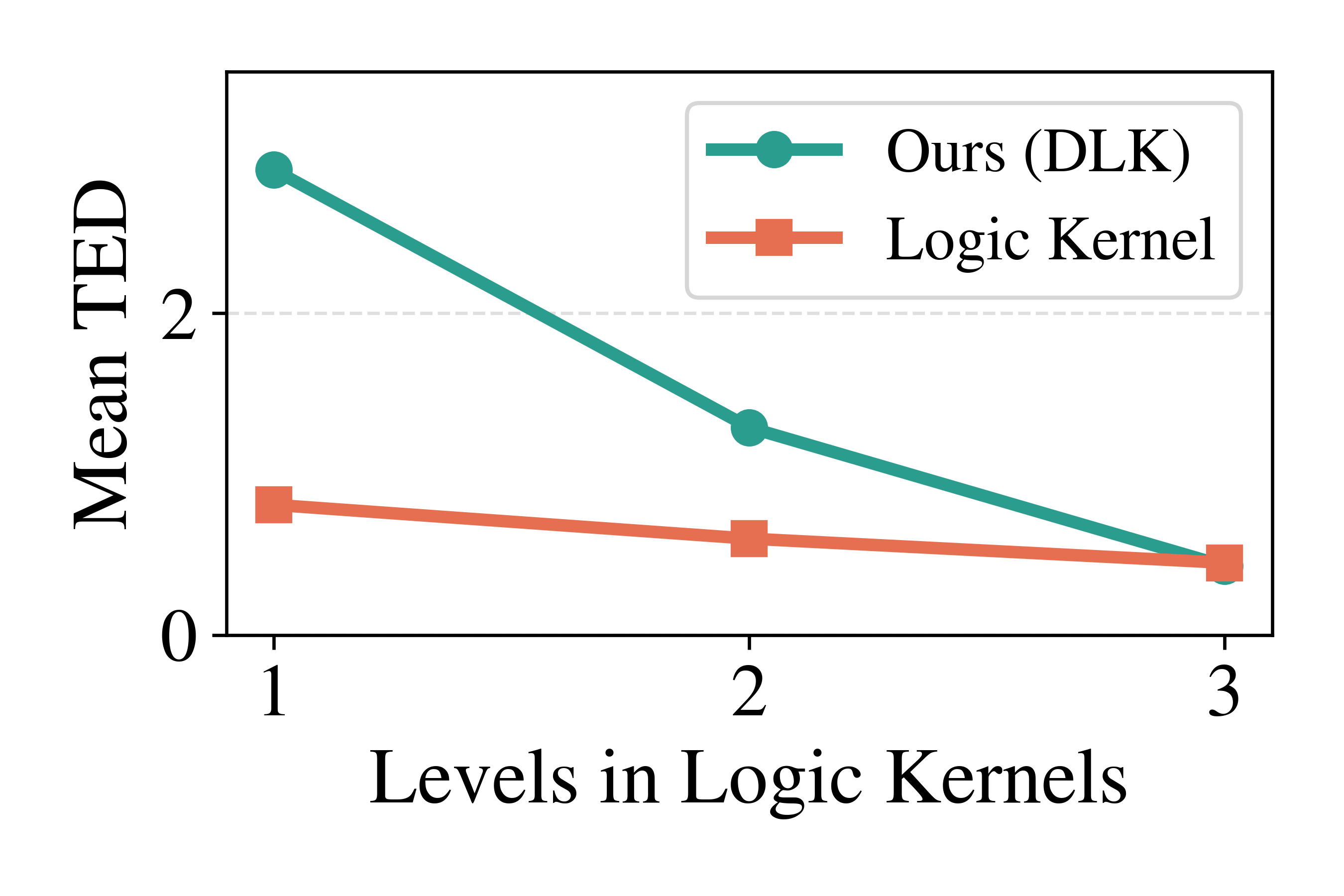}
    \caption{Level-wise decomposition of kernel diversity. }
    \label{fig:depth-ted}
    \vspace{-15pt}
\end{wrapfigure}
\paragraph{Effect of dynamic logic kernel.}
To evaluate DLK under width scaling, we fix the per-group width at $K=128$ and increase the number of groups $G$, so the total candidate width grows. As shown in Table~\ref{tab:TED_width}, DLK maintains high kernel diversity, whereas the conventional baseline exhibits low diversity that further decreases with matched width. Consistently, classification performance also improves with diversity (Figure~\ref{fig:width-scaling}), indicating that enhanced kernel diversity enables more effective utilization of model capacity.



To further investigate where DLK increases kernel diversity relative to the static baseline, we decompose pairwise TED across the logic kernel hierarchy. Specifically, we compute the pairwise mean TED at each gate level. As shown in Figure~\ref{fig:depth-ted}, the diversity gap between DLK and the static Logic Kernel baseline is largest at the first gate level, while later levels show much smaller differences. Since low diversity serves as evidence of kernel redundancy, this suggests that the primary redundancy bottleneck lies at the first gate level. A similar trend is also observed under per-gate normalization in Appendix~\ref{app:ted}.


\subsection{Early-stage Dynamic Logic Kernels}
\label{sec:edlk}
Logic gate networks can be viewed as circuits composed of interconnected logic gates. In this view, employing multiple logic kernel groups increases the diversity of available circuit patterns; however, uniformly applying these groups can lead to inefficient use of parameters and hardware resources.

Since the level-wise analysis indicates that the main bottleneck is most evident at the first gate level, adaptation can be concentrated on this early stage. Based on this insight, we propose an early-stage dynamic logic kernel that introduces level-selective dynamic routing for logic gate networks. Instead of routing entire logic kernels, EDLK dynamically selects only the first-level Boolean gate groups, exploiting the gate-level hierarchy of logic kernels for efficient utilization.

\paragraph{Method.}
As illustrated in Figure~\ref{fig:method_overview}(c), we instantiate level-selective routing by selecting only among the first-level components of the logic kernels. 
In particular, we use the same routing mechanism as in Section~\ref{sec:Selective} to compute the selection scores. 
Specifically, let $T^{(1)}_{g,k}(\cdot)$ denote the first-level components of the $k$-th kernel in the $g$-th group. We aggregate the first-level components within each group as
\begin{equation}
\mathcal{T}^{(1)}_g(\cdot) = \big[ T^{(1)}_{g,1}(\cdot) \; ; \; \cdots \; ; \; T^{(1)}_{g,K}(\cdot) \big].
\end{equation}
The router produces selection scores for these first-level candidates. During training, the adapted first-level output $\bm{Y}^{(1)}$ is computed as
\begin{equation}
\bm{Y}^{(1)}=\sum_{g=1}^{G}\alpha_{g}\, \mathcal{T}^{(1)}_{g}(\bm{X}),
\end{equation}
where $\alpha_{g}$ is defined analogously to Eq.~\eqref{eq:alpha}.

The selected first-level output is then passed through the remaining levels of the logic kernel, which are shared across groups. This yields a dynamic kernel layer designed specifically for logic gate networks. It performs input-dependent routing through the explicit Boolean gate hierarchy of logic kernels, rather than by selecting among whole real-valued filters.

\section{Experiments}
\label{experiments}
\subsection{Experimental Setup}
\label{exp_setting}

We conduct experiments on CIFAR-10~\citep{krizhevsky2009learning}, CIFAR-100~\citep{krizhevsky2009learning}, MNIST~\citep{lecun2002gradient}, Flowers-102~\citep{nilsback2008automated}, SVHN~\citep{netzer2011reading}, and Tiny-ImageNet~\citep{deng2009imagenet}, with CIFAR-10 as the primary benchmark. We report Top-1 accuracy; for width-scaling experiments, results are averaged over ten random seeds with standard-deviation error bars. All experiments use multi-threshold input binarization following prior work~\citep{petersen2022deep} and are conducted on a single NVIDIA RTX 4090 GPU. Details are provided in Appendix~\ref{app:Experimental_details}.

\begin{figure}[t]
\centering
\includegraphics[width=0.98\linewidth]{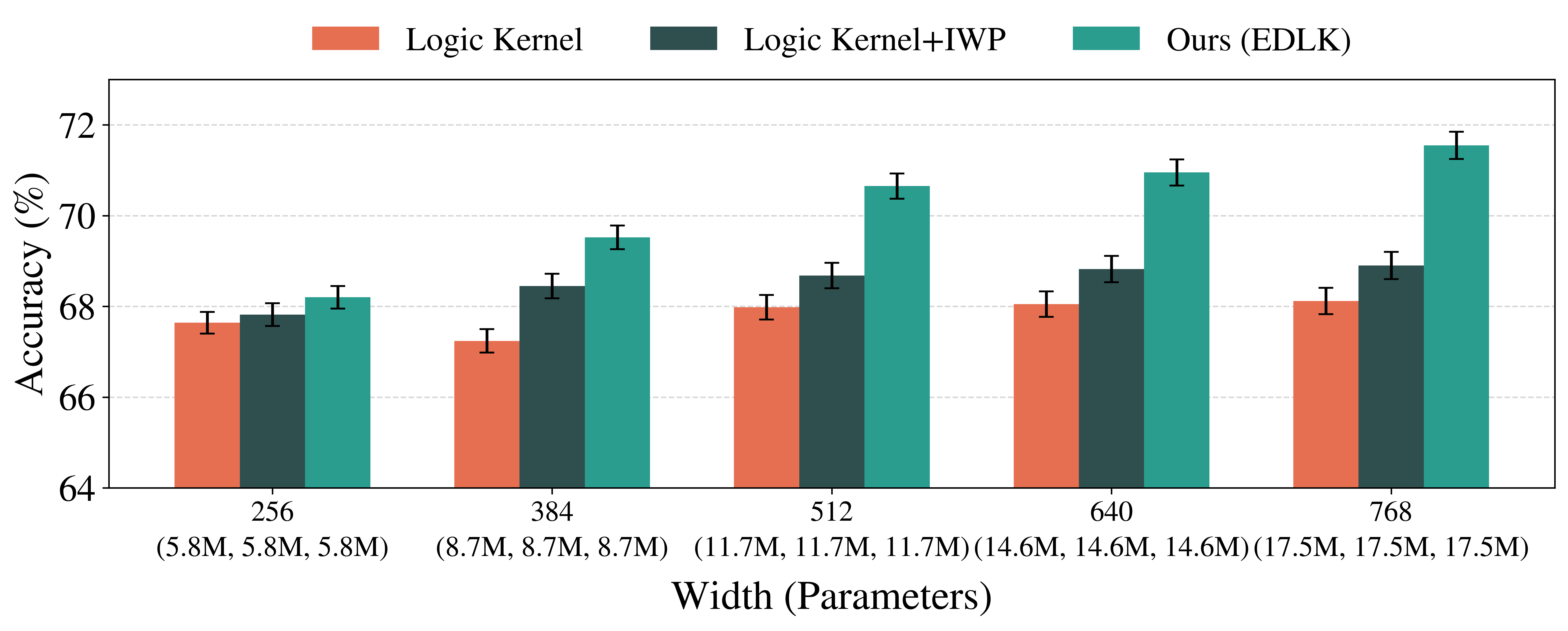}
\caption{
Top-1 accuracy on CIFAR-10 under width scaling with per-group base width $K=128$.}
\label{fig:cifar10_width2}
\end{figure}

We compare the proposed method with logic kernel baselines, including Logic Kernel~\citep{petersen2024convolutional} and Logic Kernel with IWP~\citep{ruttgers2026light}, under the same overall architecture, differing only in kernel utilization. Unless otherwise specified, our method refers to EDLK. At inference, each dynamic layer selects only one group via hard routing, as in Eq.~\eqref{eq:hard_selection}. For a fair comparison, reported parameter counts refer to the parameters used at inference time, including all candidate groups and routers. 


In the width scaling experiments, ``Width'' is defined as the total number of logic kernels in the first convolutional logic layer. Static and dynamic models are matched in total width: the static baseline uses $K \cdot G$ kernels, whereas dynamic variants use $G$ groups with $K$ kernels each. For example, in Figure~\ref{fig:cifar10_width2}, we fix $K=128$ and vary $G$, so increasing width increases both the number of groups and the total number of candidate kernels under the matched total width. 

\subsection{Results and Analysis}

\paragraph{Width scaling performance.}

We evaluate the proposed method on CIFAR-10 under different width settings. We fix the per-group base width to $K=128$ and scale the total width by increasing the number of groups $G$ under the matched-width protocol. Figure~\ref{fig:cifar10_width2} reports Top-1 accuracy averaged over ten seeds, with parameter counts shown alongside the width labels to relate performance trends to model size. As the total width increases, conventional logic kernel-based methods exhibit diminishing returns and tend to saturate, whereas EDLK continues to benefit from additional width. Consequently, the performance gap between EDLK and the static baselines becomes more pronounced at larger widths. These results indicate that simply increasing the number of logic kernels does not necessarily translate into effective model capacity, while input-dependent kernel utilization enables EDLK to make better use of the additional kernels introduced by width scaling.

\begin{table}[htbp]
\vspace{-10pt}
\centering
\caption{Top-1 accuracy across classification benchmarks. The best scores are in bold.}
\label{tab:classification_benchmark}

\begingroup
\small
\setlength{\tabcolsep}{7.0pt}

\begin{tabular}{lccccc}
\toprule
Method & MNIST & SVHN & CIFAR-100 & Flowers-102 & Tiny-ImageNet\\
\midrule
Logic Kernel
    & 98.55 & 75.58 & 58.44 & 32.40 & 10.55 \\

Logic Kernel+IWP
    & 98.67 & 76.05 & 60.33 & 33.05 & 10.86 \\

Ours (EDLK)
    & \textbf{99.10}
    & \textbf{77.00}
    & \textbf{62.46}
    & \textbf{34.26}
    & \textbf{13.35} \\
\bottomrule
\end{tabular}
\endgroup
\vspace{-5pt}
\end{table}

\paragraph{Classification benchmark results.}
Table~\ref{tab:classification_benchmark} reports Top-1 accuracy across the classification benchmarks. For a fair comparison, all methods are evaluated at a matched total width of 576, with $(K,G)=(192,3)$ for the grouped methods, while the number of input bits is set separately for each dataset. Overall, EDLK achieves the best performance across all reported datasets. While the static Logic Kernel and IWP baselines show limited gains at this larger width, EDLK continues to benefit from the increased capacity, indicating more effective utilization of the additional logic kernels.



\begin{figure}[htbp]
    \centering

    \begin{minipage}[t]{0.54\linewidth}
        \vspace{0pt}
        \centering

        \captionof{table}{Frozen versus learned routing under the matched-width setting used in Table~\ref{tab:classification_benchmark}.}

        \label{tab:random_routing}
        \vspace{3pt}

        \small
        \renewcommand{\arraystretch}{1.15}
        \setlength{\tabcolsep}{7pt}

        \begin{tabular}{lcc}
            \toprule
            Routing strategy & CIFAR-10 & Flowers-102 \\
            \midrule
            Static, no routing & 68.56 & 32.40 \\
            Frozen random      & 69.35 & 33.05 \\
            Learned (EDLK)     & \textbf{70.28} & \textbf{34.26} \\
            \bottomrule
        \end{tabular}
    \end{minipage}
    \hfill
    \begin{minipage}[t]{0.41\linewidth}
        \vspace{3pt}
        \centering

        \includegraphics[width=1.0\linewidth]{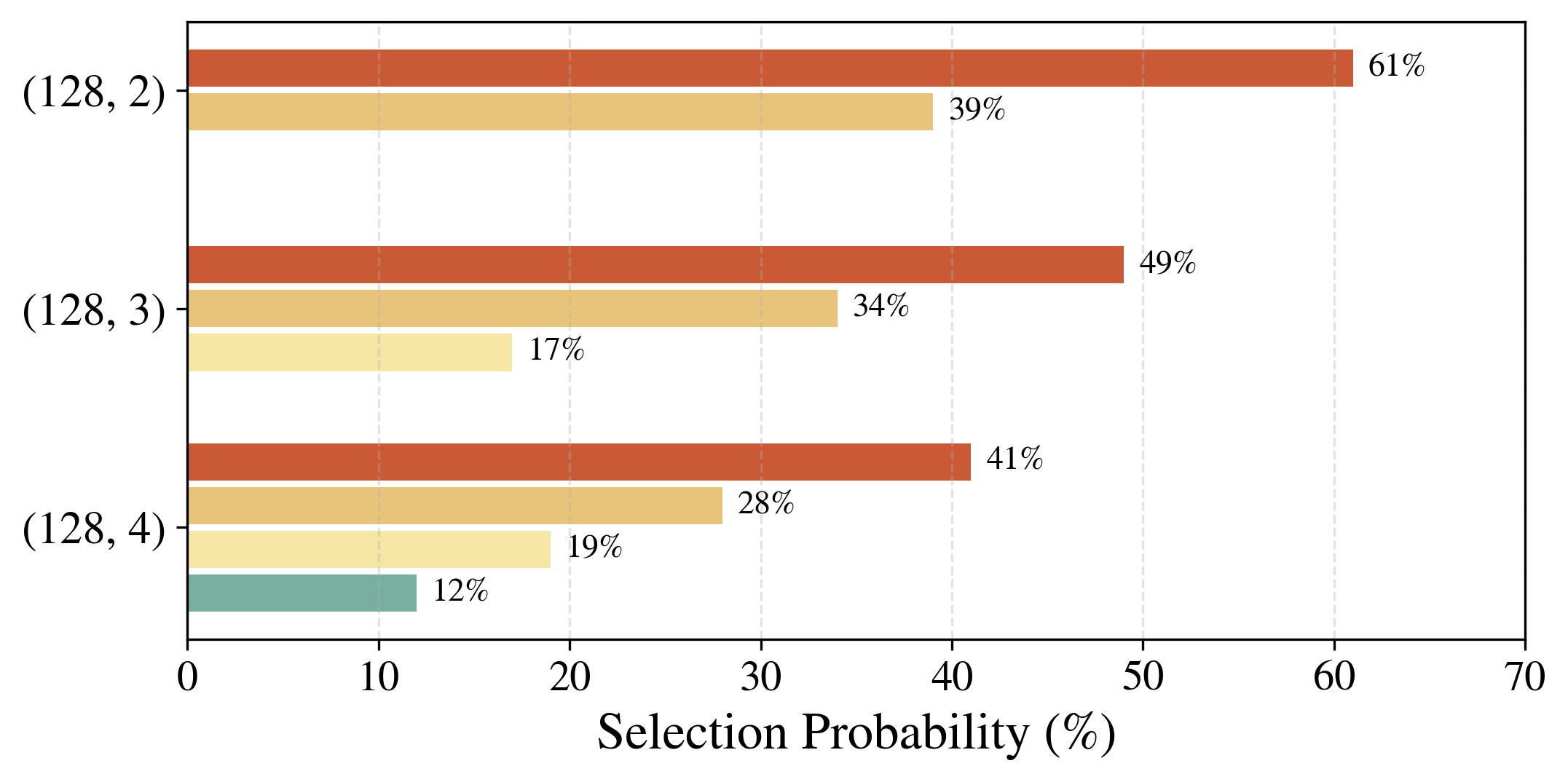}
        \vspace{-4pt}

        \captionsetup{font=footnotesize}
        \vspace{-10pt}
        \captionof{figure}{Routing distribution across kernel groups.}
        \label{fig:routing_analysis}
    \end{minipage}

    \vspace{-8pt}
\end{figure}

\paragraph{Routing behavior.}
To verify that the performance gain arises from learned input-dependent routing rather than merely introducing multiple kernel groups, we compare EDLK with an identically structured router whose parameters are randomly initialized and frozen throughout training. The candidate kernel groups and the remaining network are trained normally.

As shown in Table~\ref{tab:random_routing}, frozen random routing provides only marginal improvements over the static baseline, whereas learned routing yields substantially larger gains on both datasets. This indicates that the improvement depends on learning a functionally meaningful routing policy rather than simply introducing random routing among multiple kernel groups.

We further analyze inference-time routing statistics at conv4, aggregated over the CIFAR-10 test set. As shown in Figure~\ref{fig:routing_analysis}, routing is neither uniform nor dominated by a single group. Together, these results suggest that the learned router avoids group collapse while enabling input-dependent specialization across kernel groups.

\paragraph{Kernel diversity comparison.}

We further examine the relationship between kernel diversity and performance by jointly considering the accuracy trend in Figure~\ref{fig:width-scaling} and the diversity results at total width $256$ in Table~\ref{tab:diversity}. Consistent with our findings, the higher kernel diversity of DLK is accompanied by improved accuracy over the static baseline. Table~\ref{tab:diversity} further shows that EDLK achieves substantially higher diversity than the static baselines while remaining comparable to DLK. This suggests that level-selective adaptation to early-stage gates largely preserves the functional diversity induced by dynamic kernel utilization. Together with the observed accuracy trend, these results support the view that reduced kernel redundancy is associated with more effective use of model capacity.

\begin{table}[t]
\centering
\small

\begin{minipage}[t]{0.48\linewidth}
\centering

\captionsetup{skip=6pt}
\caption{Comparison of kernel diversity across methods at fixed width (CIFAR-10, width $256$).}
\label{tab:diversity}

\begingroup
\setlength{\tabcolsep}{10pt}
\begin{tabular}{l|c}
\toprule
Method & Diversity \\
\midrule
Logic Kernel & 2.50 \\
Logic Kernel+IWP & 3.35 \\
Ours (DLK) & 4.61 \\
Ours (EDLK) & 4.33 \\
\bottomrule
\end{tabular}
\endgroup

\end{minipage}
\hfill
\begin{minipage}[t]{0.50\linewidth}
\centering

\captionsetup{skip=6pt}
\caption{Relative reduction in logic-kernel parameters at conv4 and accuracy difference of EDLK compared to DLK.}
\label{tab:ablation_param}

\begingroup
\setlength{\tabcolsep}{8pt}
\begin{tabular}{c c c}
\toprule
$(K,G)$ & Reduction (\%) & $\Delta$ Acc (\%) \\
\midrule
(128, 2) & 21.43 & $-0.10$ \\
(128, 3) & 28.57 & $-0.03$ \\
(128, 4) & 32.14 & $-0.15$ \\
\bottomrule
\end{tabular}
\endgroup

\end{minipage}

\end{table}

\paragraph{Comparison with DLK.}
To evaluate the effectiveness of the proposed early-stage adaptation, we compare EDLK with DLK in terms of parameter efficiency and accuracy. Table~\ref{tab:ablation_param} reports the relative parameter reduction at conv4 and the corresponding Top-1 accuracy difference under matched $(K,G)$ settings. Across different $(K,G)$ settings, EDLK reduces parameters while maintaining accuracy comparable to DLK.

\begin{table}[htbp]
\vspace{-5pt}
\centering
\caption{Hardware cost and classification performance on MNIST.}
\label{tab:hardware_cost}

\begingroup
\small
\setlength{\tabcolsep}{6.0pt}

\begin{tabular}{lccccc}
\toprule
Method
& Latency (ns)$\downarrow$
& Acc. (\%)$\uparrow$
& Params
& Energy (pJ)$\downarrow$
& Area ($\mathrm{nm}^2$)$\downarrow$ \\
\midrule
Logic Kernel
& 22.06 & 95.84 & 21K & 4.15 & 12.55 \\

Logic Kernel
& 28.43 & 96.11 & 42K & 5.34 & 16.15 \\

Ours (EDLK, $G=2$)
& 23.64 & \textbf{97.29} & 22K & 4.37 & 13.21 \\
\bottomrule
\end{tabular}
\endgroup
\vspace{-8pt}
\end{table}

\paragraph{Hardware efficiency.}
To evaluate the practical overhead of EDLK beyond total parameter count, we measure FPGA latency on an Intel Cyclone V C9 device and estimate energy consumption and hardware area through synthesis using TSMC N5 technology, following prior hardware evaluation protocols~\citep{neseem2024pikelpn}. All measurements are conducted on MNIST with binary inputs. Importantly, the reported hardware costs include the entire EDLK circuit, including all candidate kernel groups, shared upper gate levels, and the complete Boolean router.

As shown in Table~\ref{tab:hardware_cost}, EDLK with 22K parameters achieves 97.29\% accuracy with 23.64\,ns latency, compared with 95.84\% and 22.06\,ns for the width-matched 21K-parameter Logic Kernel. Despite the additional routing mechanism, the resulting energy and area overhead remains modest. Moreover, compared with the wider 42K-parameter Logic Kernel, EDLK achieves higher accuracy while requiring lower latency, energy, and area. These results indicate that EDLK improves kernel utilization without sacrificing the hardware-efficient characteristics of logic gate networks.

\begin{table}[htbp]
\vspace{-8pt}
    \centering
    \caption{Hardware comparison with recent efficient models on MNIST.}
    \label{tab:hardware_comparison}
    \small
    \setlength{\tabcolsep}{5pt}
    \renewcommand{\arraystretch}{1.12}
    \begin{tabular}{lcccc}
        \toprule
        Method & BOPs$\downarrow$ & Energy (pJ)$\downarrow$ &
        Area (nm$^2$)$\downarrow$ & Acc. (\%)$\uparrow$ \\
        \midrule
        TTNet~\citep{benamira2022scalable}        & 360K & 225   & 680.75 & 98.02 \\
        DWN~\citep{pmlr-v235-bacellar24a}            & 81K  & 50.63 & 153.17 & 98.20 \\
        NeuralLUT~\citep{andronic2025neuralut}& 75K  & 46.88 & 141.82 & 97.72 \\
        \textbf{Ours (EDLK, $G=4$)}
                                  & \textbf{11K}
                                  & \textbf{6.88}
                                  & \textbf{20.80}
                                  & \textbf{98.25} \\
        \bottomrule
    \end{tabular}
    \vspace{-8pt}
\end{table}

\paragraph{Comparison with efficient models.}
We further compare EDLK with lightweight variants of recent efficiency-oriented models on MNIST in terms of accuracy and hardware cost. As shown in Table~\ref{tab:hardware_comparison}, EDLK achieves higher classification accuracy while requiring substantially fewer BOPs, lower energy consumption, and smaller hardware area than the compared methods. These results indicate that EDLK provides a favorable accuracy--efficiency trade-off while retaining the lightweight characteristics of logic gate networks.

\begin{wrapfigure}{r}{0.45\linewidth}
    \centering
    \includegraphics[width=\linewidth]{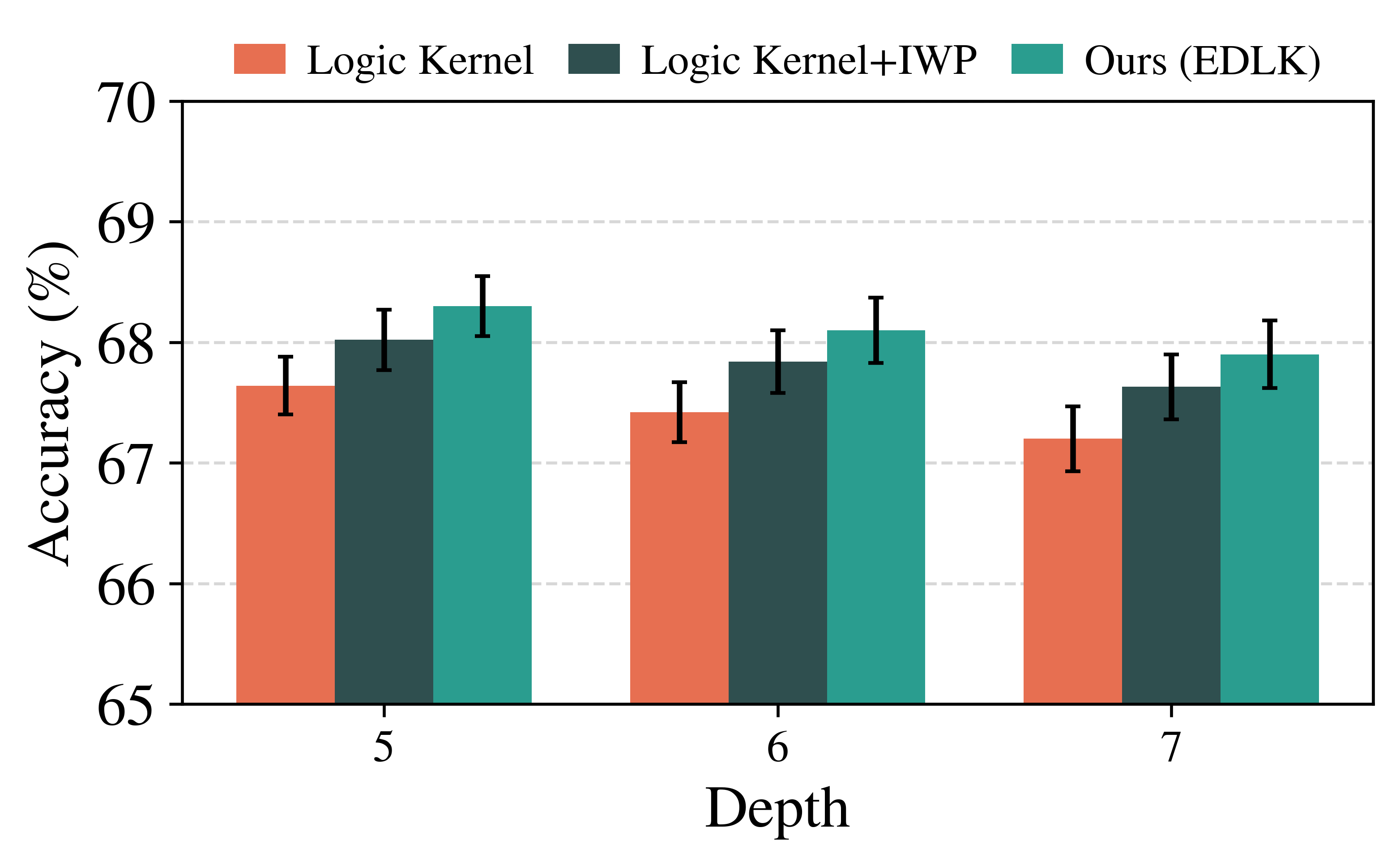}
    \vspace{-15pt}
    \caption{Top-1 accuracy under network depth scaling on CIFAR-10 with fixed width.}
    \label{fig:cifar10_depth_width}
    \vspace{-10pt}
\end{wrapfigure}

\paragraph{Effect of network depth scaling.}
We further evaluate the effect of increasing network depth under a fixed width setting, $(K,G)=(128,2)$. As shown in Figure~\ref{fig:cifar10_depth_width}, EDLK consistently outperforms the static baselines across different depth settings. However, increasing the depth does not yield additional performance gains; instead, accuracy gradually saturates or slightly decreases for all methods. This suggests that the proposed dynamic kernel utilization primarily addresses the limitations of width scaling, while the optimization and discretization challenges associated with deeper logic gate networks remain~\citep{yousefi2026mind}.



\section{Limitations}
\label{limitation}
Our study has several limitations. First, although our method improves kernel utilization under width scaling, extending the gate-based formulation to substantially larger-scale architectures, such as LLM-scale models, remains challenging. Second, our evaluation mainly focuses on image classification, leaving dense prediction tasks such as image restoration, semantic segmentation, and optical flow unexplored. Finally, logic gate networks remain an emerging paradigm and still lag behind mature architectures such as CNNs and Transformers in terms of predictive accuracy. Nevertheless, their discrete computation and hardware-friendly inference characteristics make them a promising direction for efficient model design.

\section{Future Work}
Kernel redundancy may arise from three complementary limitations in current logic gate networks: (1) \textbf{optimization}, where unstable or imbalanced training can cause different kernels to converge to similar Boolean functions; (2) \textbf{architecture}, where sparse and local information propagation may limit effective receptive-field utilization; and (3) \textbf{representation}, where fixed binary input encoding may restrict the information available to subsequent logic computations. Future work could therefore investigate more stable training strategies, improved input representations, and new architectural designs that better exploit the spatial structure of visual data, toward more scalable differentiable logic gate networks for vision tasks.

\section{Conclusion}
In this work, we revisit width scaling in logic gate networks, a fundamental yet underexplored aspect of model scaling. While scaling has been a central principle in deep learning, its behavior in logic gate networks has not been systematically understood. Through a comprehensive analysis, we reveal that naive width expansion leads to kernel redundancy, resulting in performance saturation rather than consistent gains. We further analyze redundancy across the logic kernel hierarchy and find that the first gate level is the primary source of redundancy. Based on this insight, we propose a dynamic logic kernel framework to better organize model capacity and introduce an early-stage adaptation strategy specific to logic gate networks. By formulating group selection as Boolean routing, our framework avoids external real-valued gating and preserves a fully logic gate-based inference path. Experimental results on standard benchmarks demonstrate that the proposed method improves accuracy while promoting more effective utilization of logic kernels. Overall, our study highlights that the limitation of scaling in logic gate networks is not merely a matter of insufficient capacity, but a more fundamental issue in how logical functions are structured and utilized.

\section*{Acknowledgments}
{This work was supported by the National Research Foundation of Korea (NRF)
grant funded by the Korea government (MSIT) (No. RS-2025-16068196), and National Research Foundation of Korea (NRF) grant funded by the Korea government (MSIT) (No. RS-2026-25608457).}

\bibliographystyle{unsrtnat}
\bibliography{refs}

@article{petersen2022deep,
  title={Deep differentiable logic gate networks},
  author={Petersen, Felix and Borgelt, Christian and Kuehne, Hilde and Deussen, Oliver},
  journal={Advances in Neural Information Processing Systems},
  volume={35},
  pages={2006--2018},
  year={2022}
}

@inproceedings{jacob2018quantization,
  title={Quantization and training of neural networks for efficient integer-arithmetic-only inference},
  author={Jacob, Benoit and Kligys, Skirmantas and Chen, Bo and Zhu, Menglong and Tang, Matthew and Howard, Andrew and Adam, Hartwig and Kalenichenko, Dmitry},
  booktitle={Proceedings of the IEEE/CVF Conference on Computer Vision and Pattern Recognition},
  pages={2704--2713},
  year={2018}
}

@article{howard2017mobilenets,
  title={{MobileNets}: Efficient convolutional neural networks for mobile vision applications},
  author={Howard, Andrew G and Zhu, Menglong and Chen, Bo and Kalenichenko, Dmitry and Wang, Weijun and Weyand, Tobias and Andreetto, Marco and Adam, Hartwig},
  journal={arXiv preprint arXiv:1704.04861},
  year={2017}
}

@article{hubara2016binarized,
  title={Binarized neural networks},
  author={Hubara, Itay and Courbariaux, Matthieu and Soudry, Daniel and El-Yaniv, Ran and Bengio, Yoshua},
  journal={Advances in Neural Information Processing Systems},
  volume={29},
  year={2016}
}

@inproceedings{rastegari2016xnor,
  title={{XNOR-Net}: Imagenet classification using binary convolutional neural networks},
  author={Rastegari, Mohammad and Ordonez, Vicente and Redmon, Joseph and Farhadi, Ali},
  booktitle={Proceedings of the European Conference on Computer Vision},
  pages={525--542},
  year={2016}
}

@inproceedings{park2025lut,
  title={{IM-LUT}: Interpolation mixing look-up tables for image super-resolution},
  author={Park, Sejin and Lee, Sangmin and Jin, Kyong Hwan and Jung, Seung-Won},
  booktitle={Proceedings of the IEEE/CVF International Conference on Computer Vision},
  pages={14317--14325},
  year={2025}
}

@inproceedings{tan2019efficientnet,
  title={{EfficientNet}: Rethinking model scaling for convolutional neural networks},
  author={Tan, Mingxing and Le, Quoc},
  booktitle={Proceedings of the International Conference on Machine Learning},
  pages={6105--6114},
  year={2019}
}

@article{zhou2016dorefa,
  title={{DoReFa-Net}: Training low bitwidth convolutional neural networks with low bitwidth gradients},
  author={Zhou, Shuchang and Wu, Yuxin and Ni, Zekun and Zhou, Xinyu and Wen, He and Zou, Yuheng},
  journal={arXiv preprint arXiv:1606.06160},
  year={2016}
}

@inproceedings{
liu2024qllm,
title={{QLLM}: Accurate and Efficient Low-Bitwidth Quantization for Large Language Models},
author={Jing Liu and Ruihao Gong and Xiuying Wei and Zhiwei Dong and Jianfei Cai and Bohan Zhuang},
booktitle={Proceedings of the International Conference on Learning Representations},
year={2024},
url={https://openreview.net/forum?id=FIplmUWdm3}
}

@article{petersen2024convolutional,
  title={Convolutional differentiable logic gate networks},
  author={Petersen, Felix and Kuehne, Hilde and Borgelt, Christian and Welzel, Julian and Ermon, Stefano},
  journal={Advances in Neural Information Processing Systems},
  volume={37},
  pages={121185--121203},
  year={2024}
}

@inproceedings{
yousefi2026mind,
title={Mind the Gap: Removing the Discretization Gap in Differentiable Logic Gate Networks},
author={Shakir Yousefi and Andreas Plesner and Till Aczel and Roger Wattenhofer},
booktitle={Proceedings of the Annual Conference on Neural Information Processing Systems},
year={2025},
url={https://openreview.net/forum?id=chYXaetMmz}
}

@inproceedings{
ruttgers2026light,
title={Light Differentiable Logic Gate Networks},
author={Lukas R{\"u}ttgers and Till Aczel and Andreas Plesner and Roger Wattenhofer},
booktitle={Proceedings of the International Conference on Learning Representations},
year={2026},
url={https://openreview.net/forum?id=EaGQ5luZtf}
}

@inproceedings{
ghorbani2022scaling,
title={Scaling Laws for Neural Machine Translation},
author={Behrooz Ghorbani and Orhan Firat and Markus Freitag and Ankur Bapna and Maxim Krikun and Xavier Garcia and Ciprian Chelba and Colin Cherry},
booktitle={Proceedings of the International Conference on Learning Representations},
year={2022},
url={https://openreview.net/forum?id=hR_SMu8cxCV}
}

@article{bello2021revisiting,
  title={Revisiting resnets: Improved training and scaling strategies},
  author={Bello, Irwan and Fedus, William and Du, Xianzhi and Cubuk, Ekin Dogus and Srinivas, Aravind and Lin, Tsung-Yi and Shlens, Jonathon and Zoph, Barret},
  journal={Advances in Neural Information Processing Systems},
  volume={34},
  pages={22614--22627},
  year={2021}
}

@inproceedings{vemuri2026logiccbms,
  title={{LogicCBMs}: Logic-Enhanced Concept-Based Learning},
  author={Vemuri, Deepika SN and Bellamkonda, Gautham and Pola, Aditya and Balasubramanian, Vineeth N},
  booktitle={Proceedings of the IEEE/CVF Winter Conference on Applications of Computer Vision},
  pages={6039--6048},
  year={2026}
}

@inproceedings{aczel2026ic,
  title={{IC-DLC}: Differentiable Logic Circuits for Hardware-Friendly Image Compression},
  author={Aczel, Till and Jenny, David F and B{\"u}hrer, Simon Jonas and Plesner, Andreas and Di Maio, Antonio and Wattenhofer, Roger},
  booktitle={Proceedings of the AAAI Workshop on Machine Learning for Wireless Communication and Networks},
   year={2026}
}

@article{zhang1992editing,
  title={On the editing distance between unordered labeled trees},
  author={Zhang, Kaizhong and Statman, Rick and Shasha, Dennis},
  journal={Information Processing Letters},
  volume={42},
  number={3},
  pages={133--139},
  year={1992},
  publisher={Elsevier}
}

@article{krizhevsky2009learning,
  title={Learning multiple layers of features from tiny images},
  author={Krizhevsky, Alex and Hinton, Geoffrey and others},
  year={2009},
  publisher={Toronto, ON, Canada}
}

@ARTICLE{lecun2002gradient,
  author={Lecun, Y. and Bottou, L. and Bengio, Y. and Haffner, P.},
  journal={Proceedings of the IEEE}, 
  title={Gradient-based learning applied to document recognition}, 
  year={1998},
  volume={86},
  number={11},
  pages={2278-2324},
  doi={10.1109/5.726791}}

@article{fedus2022switch,
  title={Switch transformers: Scaling to trillion parameter models with simple and efficient sparsity},
  author={Fedus, William and Zoph, Barret and Shazeer, Noam},
  journal={Journal of Machine Learning Research},
  volume={23},
  number={120},
  pages={1--39},
  year={2022}
}

@inproceedings{deng2009imagenet,
  title={Imagenet: A large-scale hierarchical image database},
  author={Deng, Jia and Dong, Wei and Socher, Richard and Li, Li-Jia and Li, Kai and Fei-Fei, Li},
  booktitle={Proceedings of the IEEE/CVF Conference on Computer Vision and Pattern Recognition},
  pages={248--255},
  year={2009}
}

@inproceedings{netzer2011reading,
  title={Reading digits in natural images with unsupervised feature learning},
  author={Netzer, Yuval and Wang, Tao and Coates, Adam and Bissacco, Alessandro and Wu, Baolin and Ng, Andrew Y and others},
  booktitle={NIPS workshop on deep learning and unsupervised feature learning},
  year={2011},
  organization={Granada}
}

@inproceedings{nilsback2008automated,
  title={Automated flower classification over a large number of classes},
  author={Nilsback, Maria-Elena and Zisserman, Andrew},
  booktitle={Proceedings of the Indian Conference on Computer Vision, Graphics \& Image processing},
  pages={722--729},
  year={2008}
}

@article{jia2016dynamic,
  title={Dynamic filter networks},
  author={Jia, Xu and De Brabandere, Bert and Tuytelaars, Tinne and Gool, Luc V},
  journal={Advances in Neural Information Processing Systems},
  volume={29},
  year={2016}
}

@inproceedings{dai2017deformable,
  title={Deformable convolutional networks},
  author={Dai, Jifeng and Qi, Haozhi and Xiong, Yuwen and Li, Yi and Zhang, Guodong and Hu, Han and Wei, Yichen},
  booktitle={Proceedings of the IEEE/CVF International Conference on Computer Vision},
  pages={764--773},
  year={2017}
}

@inproceedings{hu2018squeeze,
  title={Squeeze-and-excitation networks},
  author={Hu, Jie and Shen, Li and Sun, Gang},
  booktitle={Proceedings of the IEEE/CVF Conference on Computer Vision and Pattern Recognition},
  pages={7132--7141},
  year={2018}
}

@inproceedings{li2019selective,
  title={Selective kernel networks},
  author={Li, Xiang and Wang, Wenhai and Hu, Xiaolin and Yang, Jian},
  booktitle={Proceedings of the IEEE/CVF Conference on Computer Vision and Pattern Recognition},
  pages={510--519},
  year={2019}
}

@article{yang2019condconv,
  title={Condconv: Conditionally parameterized convolutions for efficient inference},
  author={Yang, Brandon and Bender, Gabriel and Le, Quoc V and Ngiam, Jiquan},
  journal={Advances in Neural Information Processing Systems},
  volume={32},
  year={2019}
}

@article{hua2019channel,
  title={Channel gating neural networks},
  author={Hua, Weizhe and Zhou, Yuan and De Sa, Christopher M and Zhang, Zhiru and Suh, G Edward},
  journal={Advances in Neural Information Processing Systems},
  volume={32},
  year={2019}
}

@inproceedings{chen2020dynamic,
  title={Dynamic convolution: Attention over convolution kernels},
  author={Chen, Yinpeng and Dai, Xiyang and Liu, Mengchen and Chen, Dongdong and Yuan, Lu and Liu, Zicheng},
  booktitle={Proceedings of the IEEE/CVF Conference on Computer Vision and Pattern Recognition},
  pages={11030--11039},
  year={2020}
}

@inproceedings{su2020dynamic,
  title={Dynamic group convolution for accelerating convolutional neural networks},
  author={Su, Zhuo and Fang, Linpu and Kang, Wenxiong and Hu, Dewen and Pietik{\"a}inen, Matti and Liu, Li},
  booktitle={Proceedings of the European Conference on Computer Vision},
  pages={138--155},
  year={2020}
}

@inproceedings{ma2020weightnet,
  title={Weightnet: Revisiting the design space of weight networks},
  author={Ma, Ningning and Zhang, Xiangyu and Huang, Jiawei and Sun, Jian},
  booktitle={Proceedings of the European Conference on Computer Vision},
  pages={776--792},
  year={2020}
}

@inproceedings{li2021involution,
  title={Involution: Inverting the inherence of convolution for visual recognition},
  author={Li, Duo and Hu, Jie and Wang, Changhu and Li, Xiangtai and She, Qi and Zhu, Lei and Zhang, Tong and Chen, Qifeng},
  booktitle={Proceedings of the IEEE/CVF Conference on Computer Vision and Pattern Recognition},
  pages={12321--12330},
  year={2021}
}

@inproceedings{lee2026logicir,
  title={LogicIR: Logic Gate Networks for Image Restoration},
  author={Lee, Hongjae and Son, Myungjun and Yu, Jaeseong and Jung, Seung-Won},
  booktitle={European Conference on Computer Vision},
  pages={394--411},
  year={2026},
  organization={Springer}
}

@inproceedings{neseem2024pikelpn,
  title={PikeLPN: Mitigating overlooked inefficiencies of low-precision neural networks},
  author={Neseem, Marina and McCullough, Conor and Hsin, Randy and Leichner, Chas and Li, Shan and Chong, In Suk and Howard, Andrew and Lew, Lukasz and Reda, Sherief and Rautio, Ville-Mikko and others},
  booktitle={2024 IEEE/CVF Conference on Computer Vision and Pattern Recognition (CVPR)},
  pages={15996--16005},
  year={2024},
  organization={IEEE}
}

@article{benamira2022scalable,
  title={A scalable, interpretable, verifiable \& differentiable logic gate convolutional neural network architecture from truth tables},
  author={Benamira, Adrien and Gu{\~A}{\v{S}}rand, Tristan and Peyrin, Thomas and Yap, Trevor and Hooi, Bryan},
  journal={arXiv preprint arXiv:2208.08609},
  year={2022}
}

@InProceedings{pmlr-v235-bacellar24a,
  title = 	 {Differentiable Weightless Neural Networks},
  author =       {Bacellar, Alan Tendler Leibel and Susskind, Zachary and Breternitz Jr, Mauricio and John, Eugene and John, Lizy Kurian and Lima, Priscila Machado Vieira and Fran\c{c}a, Felipe M.G.},
  booktitle = 	 {Proceedings of the 41st International Conference on Machine Learning},
  pages = 	 {2277--2295},
  year = 	 {2024},
  editor = 	 {Salakhutdinov, Ruslan and Kolter, Zico and Heller, Katherine and Weller, Adrian and Oliver, Nuria and Scarlett, Jonathan and Berkenkamp, Felix},
  volume = 	 {235},
  series = 	 {Proceedings of Machine Learning Research},
  month = 	 {21--27 Jul},
  publisher =    {PMLR},
  url = 	 {https://proceedings.mlr.press/v235/bacellar24a.html},
  
}

@inproceedings{andronic2025neuralut,
  title={Neuralut-assemble: Hardware-aware assembling of sub-neural networks for efficient lut inference},
  author={Andronic, Marta and Constantinides, George A},
  booktitle={2025 IEEE 33rd Annual International Symposium on Field-Programmable Custom Computing Machines (FCCM)},
  pages={208--216},
  year={2025},
  organization={IEEE}
}

\newpage
\appendix

\section{Experimental details and additional results}
\label{app:Experimental_details}
\subsection{Network architectures}

We use a common design principle across datasets and adapt the convolutional logic backbone according to the input resolution and input bit-depth~\citep{petersen2024convolutional, ruttgers2026light}. For example, $32\times32$ RGB datasets use the same four-stage backbone schedule, $64\times64$ RGB datasets use the same five-stage schedule, and MNIST uses a separate three-stage grayscale backbone. Throughout Appendix~\ref{app:Experimental_details}, we follow the width convention used in Section~\ref{exp_setting}.

\paragraph{Width notation.}
Specifically, $K$ denotes the base number of logic kernels in each candidate group, and $G$ denotes the number of candidate groups instantiated in a dynamic convolutional logic layer. The matched width scale $W=KG$ is defined as the candidate-kernel width of the first convolutional stage, i.e., the number of logic kernels in conv1 under the matched-width notation. Later convolutional stages are specified by fixed multipliers of this first-stage width: a stage with multiplier $m$ contains $mW=mKG$ candidate logic kernels before routing. These kernels are organized as $G$ groups of $mK$ kernels for dynamic variants, or as $mW$ kernels for static baselines. Unless otherwise stated, all convolutional stage widths below are expressed relative to this first-stage matched width $W$.

\paragraph{Static and dynamic convolutional stages.}
Static logic kernel baselines, including Logic Kernel and Logic Kernel with IWP~\citep{petersen2024convolutional, ruttgers2026light}, implement each $W$-scaled convolutional stage as a standard logic kernel layer, without input-dependent routing. Dynamic variants retain the same $W$-scaled convolutional stages, but partition their kernels into candidate groups routed on a per-input basis. The router combines the candidate-group outputs along the group dimension, rather than concatenating them into additional downstream channels.

\paragraph{Parameter-budget matching.}
Our main comparison is designed to evaluate how different kernel-utilization mechanisms perform under a comparable overall parameter budget. To avoid method-specific changes in downstream classifier capacity, we keep the subsequent standard Logic Layers fixed under the same $W$-scaled setting used by the existing baselines. Thus, the convolutional layers differ in how kernels are utilized, while the classifier head uses the same structure across methods. We report the total inference-time parameter count for each model, including all candidate kernel groups in the dynamic variants, the dynamic routers, and the subsequent standard Logic Layers.

\paragraph{Dynamic-stage selection.}
Across datasets, we instantiate dynamic logic kernels only in the highest-width convolutional stages, while keeping the rest of the convolutional logic backbone unchanged. This placement reflects the goal of DLK/EDLK: rather than redesigning the entire network, we attach dynamic routing to the stages where the candidate-kernel budget is largest under width scaling. The earlier stages therefore remain standard convolutional logic layers, and the proposed mechanism is introduced as a stage-level plug-in to the existing backbone. This allows the dynamic variants to largely retain the original architecture, while improving the utilization of widened logic kernels through only a small number of high-width stage replacements.

\subsubsection{Tiny-ImageNet}

For Tiny-ImageNet, which consists of $64 \times 64$ RGB images, we use a five-stage convolutional logic backbone. Each image is resized to $64 \times 64$ and converted into a binary representation using channel-wise multi-threshold thermometer encoding. In the reported setting, we use $C_b=127$ thresholds per RGB channel, yielding $3C_b=381$ binary input channels. The five convolutional stages are configured with widths $W$, $4W$, $16W$, $32W$, and $32W$, respectively. Each stage uses a logic tree kernel of depth $d=3$ with a $3 \times 3$ receptive field and padding of one, followed by $2 \times 2$ OR-pooling with stride $2$.

In the proposed dynamic variants, conv4 and conv5 are replaced with dynamic logic kernel layers, while conv1--conv3 remain standard convolutional logic layers. The static logic kernel baseline uses the same $W$-scaled convolutional stage structure, but implements all five stages as standard convolutional logic layers. The IWP baseline preserves the same overall topology and replaces the convolutional and fully connected logic layers with their IWP counterparts. 

\subsubsection{Flowers-102}

For Flowers-102~\citep{nilsback2008automated}, which contains variable-resolution RGB images, we use a five-stage convolutional logic backbone evaluated at $64 \times 64$ resolution. Each image is resized and cropped to $64 \times 64$ before being converted into a binary representation using thermometer encoding. We use $C_b=5$ thresholds per RGB channel, yielding $3C_b=15$ binary input channels. The five convolutional stages are configured with widths $W$, $4W$, $16W$, $32W$, and $32W$, respectively. Each stage uses a logic tree kernel of depth $d=3$ with a $3 \times 3$ receptive field and padding of one, followed by $2 \times 2$ OR-pooling with stride $2$.

In the proposed dynamic variants, conv4 and conv5 are replaced with dynamic logic kernel layers, while conv1--conv3 remain standard convolutional logic layers. The static logic kernel baseline uses the same $W$-scaled convolutional stage structure, but implements all five stages as standard convolutional logic layers. The IWP baseline preserves the same overall topology and replaces the convolutional and fully connected logic layers with their IWP counterparts. 

\subsubsection{CIFAR-100}

For CIFAR-100, which consists of $32 \times 32$ RGB images, we use a four-stage convolutional logic backbone. Each image is converted into a binary representation using channel-wise multi-threshold thermometer encoding. We use $C_b = 63$ thresholds per RGB channel, yielding $3C_b = 189$ binary input channels. The four convolutional stages are configured with widths $W$, $4W$, $16W$, and $32W$, respectively. Each stage uses a logic tree kernel of depth $d=3$ with a $3 \times 3$ receptive field and padding of one, followed by $2 \times 2$ OR-pooling with stride $2$. Following~\citep{petersen2024convolutional}, we additionally incorporate fixed low-level feature detectors to enrich the binary input representation with local spatial information.

In the proposed dynamic variants, conv4 is replaced with a dynamic logic kernel layer, while conv1--conv3 remain standard convolutional logic layers. The static logic kernel baseline uses the same $W$-scaled convolutional stage structure, but implements all four stages as standard convolutional logic layers. The IWP baseline preserves the same overall topology and replaces the convolutional and fully connected logic layers with their IWP counterparts. 

\subsubsection{CIFAR-10}

For CIFAR-10, which consists of $32 \times 32$ RGB images, we use a four-stage convolutional logic backbone. Each image is converted into a binary representation using channel-wise multi-threshold thermometer encoding. We use $C_b = 31$ thresholds per RGB channel, yielding $3C_b = 93$ binary input channels. The four convolutional stages are configured with widths $W$, $4W$, $16W$, and $32W$, respectively. Each stage uses a logic tree kernel of depth $d=3$ with a $3 \times 3$ receptive field and padding of one, followed by $2 \times 2$ OR-pooling with stride $2$.

In the proposed dynamic variants, conv4 is replaced with a dynamic logic kernel layer, while conv1--conv3 remain standard convolutional logic layers. The static logic kernel baseline uses the same $W$-scaled convolutional stage structure, but implements all four stages as standard convolutional logic layers. The IWP baseline preserves the same overall topology and replaces the convolutional and fully connected logic layers with their IWP counterparts. 

\subsubsection{SVHN}

For SVHN~\citep{netzer2011reading}, which consists of $32 \times 32$ RGB images, we use the same four-stage convolutional logic backbone as in CIFAR-10. Each image is converted into a binary representation using $C_b=2$ thresholds per RGB channel, yielding $3C_b=6$ binary input channels. The four convolutional stages are configured with widths $W$, $4W$, $16W$, and $32W$, respectively. Each stage uses a logic tree kernel of depth $d=3$ with a $3 \times 3$ receptive field and padding of one, followed by $2 \times 2$ OR-pooling with stride $2$.

In the proposed dynamic variants, conv4 is replaced with a dynamic logic kernel layer, while conv1--conv3 remain standard convolutional logic layers. The static logic kernel baseline uses the same $W$-scaled convolutional stage structure, but implements all four stages as standard convolutional logic layers. The IWP baseline preserves the same overall topology and replaces the convolutional and fully connected logic layers with their IWP counterparts. 

\subsubsection{MNIST}

For MNIST, which consists of $28 \times 28$ grayscale images, we use a shallower three-stage convolutional logic backbone. Each image is converted into a binary representation using threshold-based thermometer encoding. We use $C_b = 1$ threshold for the grayscale channel, yielding one binary input channel. The first convolutional stage uses width $W$ with a $3 \times 3$ receptive field and zero padding, followed by $2 \times 2$ OR-pooling with stride $2$. The next two convolutional stages use widths $3W$ and $9W$, respectively, both with $3 \times 3$ receptive fields, padding of one, and the same OR-pooling operation.

In the proposed dynamic variant, only the final convolutional stage is replaced with a dynamic logic kernel layer, while the first two stages remain standard convolutional logic layers. The static logic kernel baseline uses the same $W$-scaled convolutional stage structure, but implements all three stages as standard convolutional logic layers. The IWP baseline preserves the same overall topology and replaces the convolutional and fully connected logic layers with their IWP counterparts. 

\subsection{Training details}
\label{app:training_details}

\subsubsection{Objective function}
\label{app:objective_function}

We train all models with a supervised classification objective. For DLK and EDLK, we additionally use a routing load-balancing loss for the layers that contain dynamic logic kernels. Consider a mini-batch of size $B$, indexed by $n=1,\dots,B$, with labels $y^{(n)}\in\{1,\dots,C_{\mathrm{cls}}\}$. Let $\bm{z}^{(n)}\in\mathbb{R}^{C_{\mathrm{cls}}}$ denote the class logits after the final GroupSum layer.

\paragraph{Cross-entropy loss.}
The main training loss is the standard cross-entropy loss:
\begin{equation}
\mathcal{L}_{\mathrm{CE}}
=
-\frac{1}{B}
\sum_{n=1}^{B}
\log
\frac{
\exp\!\left(z^{(n)}_{y^{(n)}}\right)
}{
\sum_{c=1}^{C_{\mathrm{cls}}}
\exp\!\left(z^{(n)}_{c}\right)
}.
\label{eq:app_ce_loss}
\end{equation}
This loss supervises the final class logits. In the implementation, it is computed by applying \texttt{CrossEntropyLoss} to the output logits and labels.

\paragraph{Layer-wise load-balancing loss.}
For DLK and EDLK, we adopt the auxiliary load-balancing loss from Switch Transformers~\citep{fedus2022switch} to discourage routing collapse. The loss is applied to the routed kernel groups in each dynamic layer and averaged across dynamic layers.

\paragraph{Overall objective.}
The final training objective is
\begin{equation}
\mathcal{L}
=
\begin{cases}
\mathcal{L}_{\mathrm{CE}},
& \text{static logic kernel baselines},\\[2mm]
\mathcal{L}_{\mathrm{CE}}
+
\lambda_{\mathrm{bal}}\mathcal{L}_{\mathrm{bal}},
& \text{DLK and EDLK}.
\end{cases}
\label{eq:app_total_objective}
\end{equation}
Here, $\lambda_{\mathrm{bal}}$ controls the strength of the load-balancing regularization.

\subsubsection{Optimization and hyperparameters}
\label{app:optimization_hyperparameters}

\paragraph{Common optimization protocol.}
Unless otherwise stated, the reported experiments are trained with AdamW for $200{,}000$ iterations. The default batch size is $64$ with an initial learning rate of $0.02$. When the batch size is changed in controlled comparisons, we scale the learning rate linearly with the batch size. For the dynamic variants, the routing bit-width is selected from $[15, 31]$ according to the number of candidate groups $G$ in each layer. All experiments are conducted on a single NVIDIA RTX 4090 GPU.

\subsection{Existing assets}
\label{app:exisiting_assets}

We use CIFAR-10~\citep{krizhevsky2009learning}, CIFAR-100~\citep{krizhevsky2009learning}, MNIST~\citep{lecun2002gradient}, Flowers-102~\citep{nilsback2008automated}, SVHN~\citep{netzer2011reading}, and Tiny-ImageNet~\citep{deng2009imagenet} only as public benchmark datasets for academic evaluation. We cite their original sources, follow the official licenses or terms of use where provided, and do not redistribute, repackage, or modify the datasets.

\section{CNN precedents for dynamic kernel utilization}
\label{app:cnn_dynamic_kernel}

\paragraph{Motivation.}
Our methods are designed for logic gate networks, but motivation is related to a broader direction in CNNs: rather than applying the same fixed convolutional computation to every input, many methods adapt filters, branches, experts, channels, or groups according to the input. This appendix clarifies the relationship between DLK/EDLK and prior CNN-based approaches to input-dependent kernel utilization. While input-dependent kernel utilization has been explored in CNNs, to the best of our knowledge, this work is the first to formulate and validate this principle in logic gate networks, where both routing and computation are implemented through discretizable Boolean logic.

\paragraph{Representative CNN methods.}
Table~\ref{tab:cnn_dynamic_kernel_groups} gives a compact grouping of representative CNN methods. To avoid overloading the table, we only list the representative methods and the main adaptive object in each group; the detailed comparison is provided in the following paragraphs.

\begin{table}[t]
\centering
\small
\setlength{\tabcolsep}{4pt}
\renewcommand{\arraystretch}{1.2}
\caption{Representative CNN methods related to adaptive kernel or capacity utilization.}
\label{tab:cnn_dynamic_kernel_groups}
\begin{tabular}{@{}p{0.24\linewidth}p{0.47\linewidth}p{0.21\linewidth}@{}}
\toprule
Group & Representative methods & Adapted object \\
\midrule
\textbf{Kernel generation and spatial adaptation}
&
\begin{tabular}[t]{@{}l@{}}
Dynamic Filter Networks~\citep{jia2016dynamic} \\
Deformable ConvNets~\citep{dai2017deformable} \\
WeightNet~\citep{ma2020weightnet} \\
Involution~\citep{li2021involution}
\end{tabular}
&
Filters, weights, sampling support, or spatial kernels \\
\addlinespace

\textbf{Kernel, branch, and expert selection}
&
\begin{tabular}[t]{@{}l@{}}
Selective Kernel Networks~\citep{li2019selective} \\
CondConv~\citep{yang2019condconv} \\
Dynamic Convolution~\citep{chen2020dynamic}
\end{tabular}
&
Kernels, branches, or experts \\
\addlinespace

\textbf{Channel and group routing}
&
\begin{tabular}[t]{@{}l@{}}
Squeeze-and-Excitation Networks~\citep{hu2018squeeze} \\
Channel Gating Neural Networks~\citep{hua2019channel} \\
Dynamic Group Convolution~\citep{su2020dynamic}
\end{tabular}
&
Channels or groups \\
\bottomrule
\end{tabular}
\end{table}

\paragraph{Kernel generation and spatial adaptation.}
The first group modifies the convolutional operator itself. Dynamic Filter Networks generate filters conditioned on the input, showing that convolutional filters do not need to be fixed after training~\citep{jia2016dynamic}. Deformable ConvNets make the spatial support of convolution adaptive by learning offsets for sampling locations~\citep{dai2017deformable}. WeightNet studies convolutional weight generation through a weight-generating network, providing a related perspective on how convolutional weights can be parameterized beyond a directly stored static tensor~\citep{ma2020weightnet}. Involution further changes the standard convolutional design by using spatial-specific operators~\citep{li2021involution}. These methods show that the effective convolutional operation can be made more flexible than a single fixed filter bank. 

\paragraph{Kernel, branch, and expert selection.}
The second group is the closest CNN analogue to DLK. Selective Kernel Networks use multiple convolutional branches with different receptive-field sizes and adaptively combine their outputs through attention~\citep{li2019selective}. CondConv learns example-dependent convolutional kernels by combining multiple expert kernels with input-conditioned routing weights~\citep{yang2019condconv}. Dynamic Convolution similarly aggregates multiple convolution kernels using attention over kernels~\citep{chen2020dynamic}. These methods share the idea that a layer can contain multiple candidate transformations and adaptively select or mix them for each input. 

\paragraph{Channel and group routing.}
The third group adapts computation at the channel or group level. Squeeze-and-Excitation Networks recalibrate channel-wise feature responses by modeling dependencies between channels~\citep{hu2018squeeze}. Channel Gating Neural Networks dynamically skip subsets of channel computations to reduce unnecessary computation~\citep{hua2019channel}. Dynamic Group Convolution selects input channels for each group conditioned on the input, improving the utilization of grouped convolution~\citep{su2020dynamic}. These methods are not direct dynamic kernel methods in the same sense as CondConv or Dynamic Convolution. However, they support the broader view that model capacity can be used more effectively when channels or groups are adaptively emphasized rather than uniformly used. 

\paragraph{DLK formulation.}
DLK follows the general principle of input-dependent kernel utilization, but reformulates it for the computational structure of logic gate networks. Unlike CNN-based dynamic kernel methods, where the adapted object is typically a real-valued convolutional filter, DLK treats each kernel as a discretizable Boolean logic tree. It constructs $G$ candidate groups of logic kernels, $\{\mathcal{T}_g\}_{g=1}^{G}$, and uses a logic-gate-based router $\mathcal{R}(\cdot)$ to produce input-dependent group scores. During training, the groups are combined through the soft routing weights in Eq.~\eqref{eq:alpha}, while inference replaces this relaxation with hard selection as in Eq.~\eqref{eq:hard_selection}. Thus, DLK performs conditional selection over Boolean logic-kernel groups rather than real-valued filter modulation, keeping both routing and computation within the logic-gate-based inference formulation.

\paragraph{Selection granularity.}
A key design choice is the level at which dynamic selection is applied. Whole-layer selection is too coarse because all kernels share the same routing decision, whereas independent gate-wise routing is too fine-grained and would weaken the shared hierarchical structure of a logic kernel. DLK therefore performs selection at the structured kernel group level, allowing different Boolean kernel groups to specialize to different input patterns while preserving a simple logic-compatible routing structure.

This granularity also motivates EDLK. A logic kernel is a hierarchical Boolean tree, so dynamic selection can be applied inside the kernel rather than only to the entire kernel. The level-wise TED analysis in Figure~\ref{fig:depth-ted} shows that the diversity gap between DLK and the static Logic Kernel baseline is largest at the input-adjacent first gate level, suggesting that the main redundancy bottleneck appears where elementary Boolean primitives are formed. Based on this observation, EDLK localizes dynamic selection to first-level gate groups and shares the remaining upper levels across groups.

\paragraph{EDLK formulation.}
EDLK applies the same routing principle as DLK, but adapts the selected units to first-level Boolean gate groups. Specifically, it constructs $G$ candidate first-level operators $\{\mathcal{T}^{(1)}_g\}_{g=1}^{G}$ and computes the routing weights $\bm{\alpha}=[\alpha_1,\ldots,\alpha_G]$ using the router in
Section~\ref{sec:Selective}. During training, the adapted first-level output is computed as
\begin{equation}
    \bm{Y}^{(1)}
    =
    \sum_{g=1}^{G}
    \alpha_g \mathcal{T}^{(1)}_g(\bm{X}),
\end{equation}
and the remaining logic-tree levels are shared across groups. At inference time, the soft aggregation is replaced by hard selection of a single first-level gate group, following the same soft-to-hard routing rule as DLK.

This design is specific to logic gate networks. Since the first level forms elementary Boolean interactions from the input activations, selecting among first-level gate groups changes the Boolean primitives supplied to the shared upper-level circuit. EDLK therefore introduces input-dependent variation where Boolean primitives are formed, while preserving the compact and compositional structure of the deeper logic gates.

\paragraph{Logic gate network-specific dynamic routing.}
DLK and EDLK formulate dynamic routing for a setting that does not naturally arise in ordinary real-valued kernels. CNN-based dynamic-kernel methods typically adapt filters, branches, channels, or experts at the tensor level. In contrast, a logic kernel is a discretizable Boolean circuit composed of explicit gate levels, where each internal node represents a finite two-input Boolean function. This structure enables a routing mechanism that is specific to logic gate networks: DLK selects among Boolean kernel groups, while EDLK selects among structured first-level gate groups inside the kernel hierarchy.

This design highlights why EDLK is specific to logic gate networks rather than a generic compression of dynamic kernels. In ordinary real-valued kernels, there is no explicit hierarchy of discrete Boolean gates where routing can be localized; adaptation is therefore typically defined over filters, channels, branches, or experts. A logic kernel, however, is a gate-level Boolean circuit, so its input-adjacent gate groups form a meaningful and routable unit: they determine the elementary Boolean primitives that are later composed by the shared upper gates. EDLK exploits exactly this logic-specific structure by applying dynamic selection only where Boolean primitives are formed, while keeping the remaining circuit shared. Thus, EDLK introduces a dynamic mechanism that is both needed for addressing first-level redundancy in logic kernels and implementable only because logic gate networks expose an explicit Boolean gate hierarchy.

\section{Tree Edit Distance}
\label{app:ted}

This appendix provides additional details on the function-aware tree edit distance (TED) used for the kernel diversity analysis in Sections~\ref{sec:understanding_logic_tree} and~\ref{sec:Selective}. Since logic kernels within the same convolutional logic layer share an identical tree topology, two kernels can be compared by aligning their internal gate nodes. Let $T_A$ and $T_B$ denote two logic kernels with $L$ internal gates, and let $g^{T_A(n)}$ and $g^{T_B(n)}$ be the Boolean operators assigned to the $n$-th aligned node of each kernel. The function-aware TED is defined as
\begin{equation}
 d(T_A,T_B)=\sum_{n=1}^{L}\delta_f\!\left(g^{T_A(n)},g^{T_B(n)}\right).
\label{eq:app_functional_ted}
\end{equation}

\paragraph{Truth-table representation.}
Each gate in a logic tree corresponds to a Boolean function defined by its outputs over the input-pair set $\mathcal{A}=\{(0,0),(0,1),(1,0),(1,1)\}$. We therefore represent a gate $g$ using the truth vector
\begin{equation}
\mathbf{b}(g)=\big[g(0,0),\,g(0,1),\,g(1,0),\,g(1,1)\big]\in\{0,1\}^{4}.
\label{eq:app_truth_vector}
\end{equation}
Table~\ref{tab:app_truth_table} summarizes the 16 possible two-input Boolean operators under this fixed ordering.

\begin{table}[t]
\centering
\small
\setlength{\tabcolsep}{4.5pt}
\caption{Truth-table representation of the 16 two-input Boolean operators.}
\label{tab:app_truth_table}
\begin{tabular}{@{}c l c c c c@{}}
\toprule
$\mathbf{b}(g)$ & Boolean operator $g(a_1,a_2)$ & $00$ & $01$ & $10$ & $11$ \\
\midrule
0000 & $0$ & 0 & 0 & 0 & 0 \\
0001 & $a_1 \wedge a_2$ & 0 & 0 & 0 & 1 \\
0010 & $a_1 \wedge \neg a_2$ & 0 & 0 & 1 & 0 \\
0011 & $a_1$ & 0 & 0 & 1 & 1 \\
0100 & $\neg a_1 \wedge a_2$ & 0 & 1 & 0 & 0 \\
0101 & $a_2$ & 0 & 1 & 0 & 1 \\
0110 & $a_1 \oplus a_2$ & 0 & 1 & 1 & 0 \\
0111 & $a_1 \vee a_2$ & 0 & 1 & 1 & 1 \\
1000 & $\neg(a_1 \vee a_2)$ & 1 & 0 & 0 & 0 \\
1001 & $\neg(a_1 \oplus a_2)$ & 1 & 0 & 0 & 1 \\
1010 & $\neg a_2$ & 1 & 0 & 1 & 0 \\
1011 & $a_1 \vee \neg a_2$ & 1 & 0 & 1 & 1 \\
1100 & $\neg a_1$ & 1 & 1 & 0 & 0 \\
1101 & $\neg a_1 \vee a_2$ & 1 & 1 & 0 & 1 \\
1110 & $\neg(a_1 \wedge a_2)$ & 1 & 1 & 1 & 0 \\
1111 & $1$ & 1 & 1 & 1 & 1 \\
\bottomrule
\end{tabular}
\end{table}

Using this truth-vector representation, the node-wise functional cost is computed as
\begin{equation}
\delta_f(g,g')=
\sum_{(a_1,a_2)\in\mathcal{A}}
\mathbf{1}\!\left(g(a_1,a_2)\neq g'(a_1,a_2)\right)
=\left\lVert\mathbf{b}(g)-\mathbf{b}(g')\right\rVert_0 .
\label{eq:app_delta_f}
\end{equation}
This cost counts the number of input cases for which two gates produce different outputs. For example, $\oplus$ and $\vee$ differ only at $(1,1)$, yielding $\delta_f(\oplus,\vee)=1$, whereas $\oplus$ and $\wedge$ differ in three input cases, yielding $\delta_f(\oplus,\wedge)=3$. Hence, unlike a binary same-or-different gate comparison, the proposed cost captures the degree of functional discrepancy between Boolean operators.

\paragraph{Kernel diversity and level-wise analysis.}
Given a set of $N$ logic kernels in a convolutional logic layer, $\mathcal{T}=\{T_1,\dots,T_N\}$, we quantify kernel diversity using the mean pairwise distance
\begin{equation}
\mathcal{D}_{\mathcal{T}}
=\frac{1}{\binom{N}{2}}\sum_{i<j}d(T_i,T_j).
\label{eq:app_kernel_diversity}
\end{equation}
The reported diversity score is obtained by averaging this quantity across the convolutional logic layers included in the analysis.

For the level-wise analysis in Figure~\ref{fig:depth-ted}, we compute the same distance while restricting the summation to gates at a specific tree level. We index levels from the input-adjacent gate level to the output-adjacent root level. Let $\mathcal{N}_{\ell}$ denote the set of aligned nodes at level $\ell$. The level-wise distance is defined as
\begin{equation}
 d^{(\ell)}(T_A,T_B)
 =\sum_{n\in\mathcal{N}_{\ell}}
 \delta_f\!\left(g^{T_A(n)},g^{T_B(n)}\right).
\label{eq:app_level_ted}
\end{equation}
The corresponding level-wise mean pairwise diversity is
\begin{equation}
\mathcal{D}_{\mathcal{T}}^{(\ell)}
=\frac{1}{\binom{N}{2}}\sum_{i<j}d^{(\ell)}(T_i,T_j).
\label{eq:app_level_diversity}
\end{equation}
This quantity gives an additive decomposition of the full-kernel TED across the hierarchy of the logic kernel. In our depth-$3$ logic kernels, the three gate levels contain different numbers of gates. Therefore, $\mathcal{D}_{\mathcal{T}}^{(\ell)}$ should be interpreted as the additive contribution of level $\ell$ to full-kernel diversity.

\begin{figure}[htbp]
    \centering
    \includegraphics[width=0.55\linewidth]{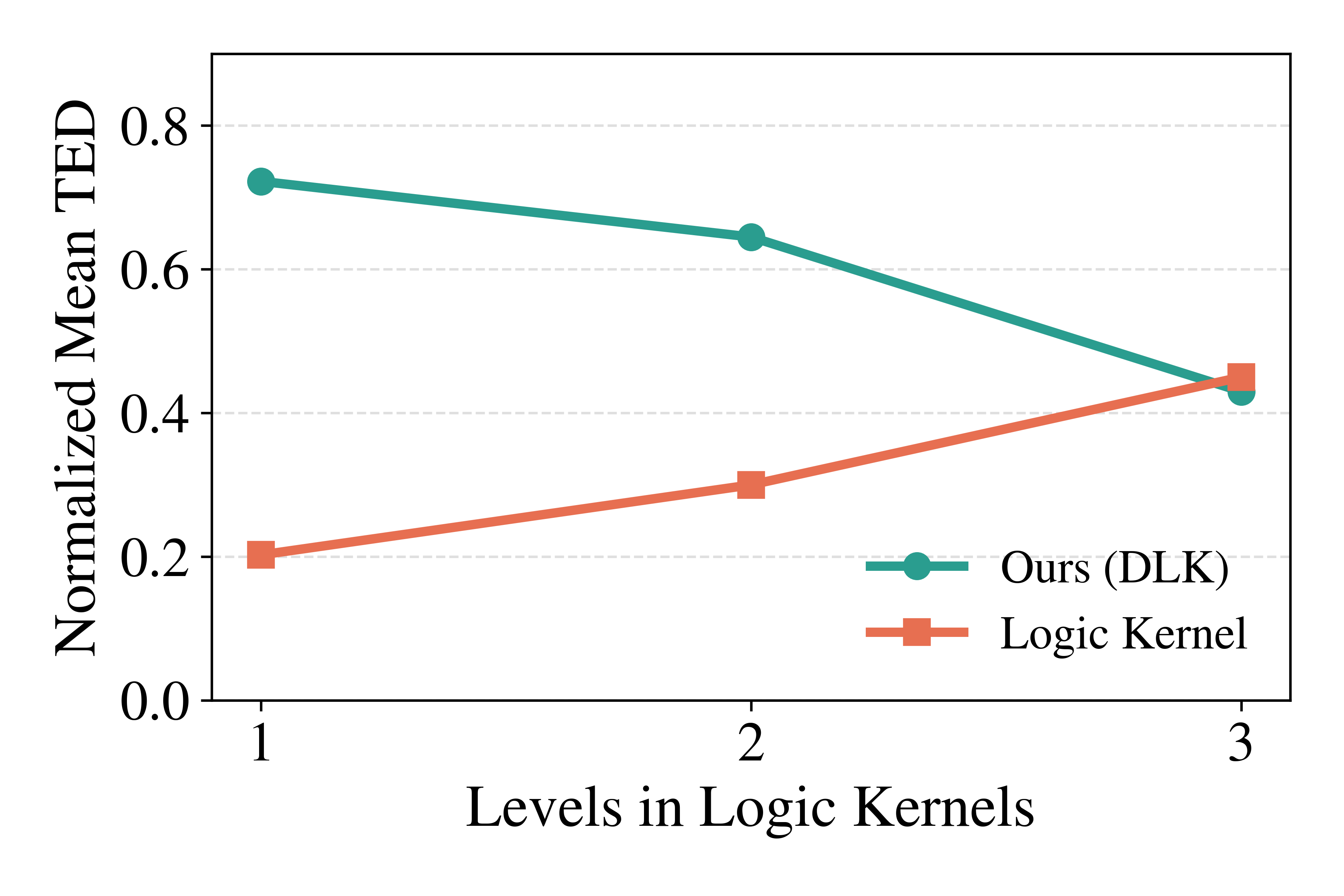}
    \caption{Per-gate normalized level-wise TED on CIFAR-10. Each level-wise TED value is divided by the number of gates at that level. A similar first-level trend is observed after normalization.}
    \label{fig:app_normed_level_ted}
\end{figure}

\paragraph{TED-based empirical results.}
Using the discretized gates obtained after training, we compute the mean pairwise function-aware TED between logic kernels within each convolutional logic layer. These layer-wise distances are then averaged across the analyzed layers to obtain the reported diversity score. For the level-wise analysis, the same computation is performed independently at each gate level, enabling a direct comparison of functional diversity across different depths of the logic tree. For the normalized analysis in Figure~\ref{fig:app_normed_level_ted}, we divide each level-wise distance by the number of gates at the corresponding level before averaging over kernel pairs.

\newpage

\end{document}